\documentclass[11pt]{article}

\usepackage[preprint]{acl}
\usepackage{times}
\usepackage{latexsym}
\usepackage{multirow}
\usepackage{booktabs}
\usepackage[T1]{fontenc}
\usepackage[utf8]{inputenc}
\usepackage{microtype}
\usepackage{inconsolata}
\usepackage{graphicx}
\usepackage{amsmath,amssymb}
\usepackage{enumitem}
\usepackage{hyperref}
\usepackage{xcolor}
\usepackage{dblfloatfix}
\usepackage{float}
\usepackage{xspace}
\usepackage{subcaption}
\usepackage{tikz}
\usepackage{algorithmic}
\usepackage{algorithm}
\usepackage{cleveref}
\usepackage[most]{tcolorbox}
\usepackage{fvextra}

\DefineVerbatimEnvironment{PromptVerbatim}{Verbatim}{
  fontsize=\footnotesize,
  breaklines=true,
  breakanywhere=false,
  breaksymbolleft={},
  breaksymbolright={}
}
\newtcolorbox{examplebox}[1]{
  breakable,
  colback=gray!2,
  colframe=gray!45,
  boxrule=0.4pt,
  arc=1mm,
  left=1.5mm,
  right=1.5mm,
  top=1mm,
  bottom=1mm,
  title={#1},
  fonttitle=\bfseries\footnotesize
}
\newcommand{\prl}{\ensuremath{\mathtt{PRL}}\xspace}
\newcommand{\wmrl}{\ensuremath{\mathtt{WMRL}}\xspace}
\newcommand{\wmsft}{\ensuremath{\mathtt{WMSFT}}\xspace}
\newcommand{\wmsftprl}{\ensuremath{\mathtt{WMSFT{+}PRL}}\xspace}
\newcommand{\wmprl}{\ensuremath{\mathtt{WMRL{+}PRL}}\xspace}

\newcommand{\upd}[1]{\Delta_{#1}}
\newcommand{\ddupd}[1]{\Delta\Delta_{#1}}
\newcommand{\mdl}[1]{M_{#1}}

\newcommand{\dprl}{\upd{\prl}}
\newcommand{\dwmrl}{\upd{\wmrl}}
\newcommand{\dwmsft}{\upd{\wmsft}}
\newcommand{\dwmprl}{\upd{\wmprl}}
\newcommand{\ddprl}{\ddupd{\prl}}

\newcommand{\erank}{\operatorname{eRank}}

\definecolor{h1bg}{HTML}{F6F8FF}
\definecolor{h1frame}{HTML}{6A7FDB}
\definecolor{h2bg}{HTML}{F4FFF7}
\definecolor{h2frame}{HTML}{4BA66A}
\definecolor{h3bg}{HTML}{FFF8F1}
\definecolor{h3frame}{HTML}{D9904A}

\newtcolorbox{hypobox}[2]{
  width=\columnwidth,
  colback=#1,
  colframe=#2,
  boxrule=0.45pt,
  arc=1.5pt,
  boxsep=2pt,
  left=3pt,
  right=3pt,
  top=3pt,
  bottom=3pt,
  before skip=3pt,
  after skip=3pt,
  fontupper=\footnotesize,
}
\usepackage{tabularx}
\usepackage{booktabs}
\usepackage{makecell}

\newcolumntype{Y}{>{\raggedright\arraybackslash}X}

\title{How do World Models and Policies Compose in LLM Agents? \\ A Joint Spectral and Behavioral Account}

\author{
  \textbf{Ruize Xu\textsuperscript{1}},
  \textbf{Xiao Yu\textsuperscript{2}},
  \textbf{Yujin Tang\textsuperscript{1}},
  \textbf{Chenming Shang\textsuperscript{1}},
  \textbf{Nikhil Singh\textsuperscript{1}}
\\
  \textsuperscript{1}Dartmouth College \quad
  \textsuperscript{2}Columbia University
\\
  \texttt{
    \{\href{ruize.xu.gr@dartmouth.edu}{ruize.xu.gr},\href{yujin.tang.gr@dartmouth.edu}{yujin.tang.gr},\href{chenming.shang.gr@dartmouth.edu}{chenming.shang.gr},\href{nikhil.u.singh@dartmouth.edu}{nikhil.u.singh}\}@dartmouth.edu
    }\\
  \texttt{ 
    \href{xy2437@columbia.edu}{xy2437@columbia.edu}
  }
}

\begin{document}
\maketitle

\begingroup
\renewcommand\thefootnote{}
\footnotetext{%
\textbf{Code: }
\href{https://github.com/Rick-Xu315/WM_Spectral}
{\texttt{github.com/Rick-Xu315/WM\_Spectral}}
}
\addtocounter{footnote}{0}
\endgroup
\begin{abstract}
How do LLM agents come to both understand environments they act in and master tasks set within them? Through controlled experiments combining world-model training (next-state prediction) and policy training (reward maximization), we investigate this question. We dissect the resulting models through their additive parameter updates. Geometrically, we find effective world-model updates are low-rank and share an \textit{input}-feature subspace with policy updates while writing to nearly orthogonal \textit{output} directions, whether trained separately or sequentially. However, we find that, in projection interventions, the sequential update induces more robustness than separate policy RL when removing the world model's leading input directions, suggesting that it has learned alternative input pathways. Behaviorally, we find the sequentially trained agent explores a wider range of states and actions. Based on this, we ask: does policy training preserve world knowledge as well as it \textit{could}? We probe this with training-free merging built on the geometrically motivated input basis plus an online world-model loss during policy RL, and show both improve over the untreated baseline. Our findings suggest world knowledge and task-directed ability can be learned in geometrically complementary forms, and that future post-training pipelines should consider how best to engineer the interface between them.
\end{abstract}

\section{Introduction}
\label{sec:intro}
We know things about the world, and we do things in it. Language models too are increasingly studied through this dual lens (e.g., their \textit{world models}~\cite{li2023emergent,toshniwal2022chess,guan2023leveraging,vafa2024evaluating}) and are even explicitly trained to internalize both these capacities. For example, post-training pipelines for LLM-based agents increasingly adopt a two-stage formula: a \emph{world-model} stage that trains the model to anticipate environment dynamics, followed by a \emph{policy} stage that optimizes end-task reward~\citep{yu2025dynathink, zhang2026agent,
yu2025dynamind, chen2025spa, yu2026rwml}. This sequential pipeline consistently improves over policy training from a base model, often by sizable margins.

What is the relationship between these two objects: knowledge and ability? Intuitively, knowing your environment well helps you select better actions. Still, the two are not the same: by analogy, radiologists don't necessarily make good surgeons. Despite the empirical success of world-model learning, the precise mechanism by which it benefits downstream policies remains opaque. For example, do world models act as a warm-up that places the policy at a better initialization in a region of weight space it would have reached on its own? Does it identify a useful subspace the policy stage then refines? Or does it leave behind a representational structure on which the policy builds in meaningfully \emph{complementary} directions?

Getting good answers to these questions is potentially consequential for both interpretation and design. If world-model learning accelerates the policy's discovery of the same solution, the second
stage should in principle be able to absorb the first, and the pipeline could be collapsed without loss. If instead the two stages contribute structurally different things, then how to preserve, combine, and schedule them is an important design lever for agentic post-training. Prior work has documented the benefit of the pipeline but rarely its mechanism. Existing analyses either evaluate the composed pipeline as a black box~\citep{yu2026rwml, chen2025spa} or characterize properties of post-training updates in isolation~\citep{aghajanyan2021intrinsic,
hu2022lora, zhu2025path}, without directly contrasting world-model and policy stages. Perhaps closest to the present work is~\citet{zhu2025path}, which uses spectral structure to show how RLVR largely preserves principal subspaces and updates off-principal directions vs. SFT. Similarly, we analyze post-training updates through spectral geometry and, consistent with their results, we find RL and SFT updates differ sharply in the world model regime. However, our primary focus is within the RL regime, analyzing the input-feature vs. output directions.

We take a first step toward understanding how world-model and policy stages interact via parameter-space dissection of a controlled family of post-trained checkpoints. For each post-training paradigm $X$, we form the additive parameter update $\Delta_X = \theta_X - \theta_0$ relative to the base model; the policy stage's incremental contribution on top of the world model can then be analyzed separately. We then conduct a series of spectral interventions (subspace projection, rank truncation, and magnitude-based pruning) and behavioral measurements (held-out next-state prediction entropy, state coverage, and patterns of success). Our results show meaningful complementarity between world-model and policy updates: the two stages share an input-feature subspace but write to nearly orthogonal output directions, and sequential training additionally equips the policy update with input-pathway redundancy that base-initialized policy training cannot acquire. We also show that the two-stage agent exhibits useful behavioral capacities (e.g., preserved world-model calibration, expanded state coverage) over the policy-only version.

In all, this paper contributes an exploratory empirical analysis of two-stage agentic post-training: a parameter-space dissection that maps out the geometric structure underlying the two-stage recipe's empirical success, across two benchmarks and two models, in which we ask:
\begin{enumerate}[label=\textbf{RQ\arabic*},ref=RQ\arabic*]
    \item Are there geometric and behavioral signatures of effective world-model training?
    \item What is the relationship between parameter updates produced by world-model and policy training, independently vs. sequentially?
    \item Is the world-model preserved enough through policy training, and can we improve this?
\end{enumerate}

\noindent
This work then contributes:
\begin{enumerate}
    \item \textbf{Empirical evidence} showing \textbf{spectral and behavioral signatures of effective world-model training} (RQ1; \S\ref{sec:spectral}, \S\ref{sec:how-helps}).
    \item A \textbf{$\Delta$-decomposition} and  left/right \textbf{subspace projection methodology}, using which we find \textbf{geometric evidence of stage complementarity} (RQ2; \S\ref{sec:relationship}). We find that RL-trained world models' left singular subspaces (output directions) are nearly orthogonal to the downstream policy's, while their right singular subspaces (input-feature directions) are strongly aligned. Sequential training additionally equips the policy update with redundant input pathways unavailable from base-only.
    \item \textbf{Design implications} for \textit{preserving} world models during policy training, realized via \textbf{merging and SFT experiments} (RQ3; \S\ref{sec:preserve}). The complementarity finding predicts that the two stages should compose constructively. We test both a training-free KnOTS/TIES merge and an online world model SFT auxiliary loss during policy learning (similar to concurrent work by \citet{shrivastava2026echo}), both of which improve over the untreated baseline.
\end{enumerate}

\section{Preliminaries}
\label{sec:prelim}

\subsection{Problem Setup}
We formulate multi-step task completion as a Markov Decision Process
$\langle \mathcal{S}, \mathcal{A}, \mathcal{T}, \mathcal{R}, \gamma \rangle$
\citep{sutton2018reinforcement}. A ReAct-style \citep{yao2023react} LLM agent
$\pi_\theta$ interleaves reasoning and acting: at step $t$, it receives an
observation $s_t \in \mathcal{S}$ together with up to $H$ turns of
interaction history $\langle s_{t-H}, a_{t-H}, \ldots, s_t \rangle$, produces
thought $\tau_t$, and emits an action
$a_t \sim \pi_\theta(\cdot \mid s_{t-H:t}, a_{t-H:t-1})$; the environment
returns the next observation $s_{t+1} \sim \mathcal{T}(\cdot \mid s_t, a_t)$.
The loop terminates with a task-success reward $r_T$. 

\paragraph{World model.}
A world model $\mathcal{W}_\phi$ is a learned surrogate for the transition
$\mathcal{T}$ \citep{ha2018world, hafner2023dreamerv3}: it predicts the next
state $\hat{s}_{t+1} \sim \mathcal{W}_\phi(\cdot \mid s_{t-H:t}, a_{t-H:t})$.
We write $\hat{s}_t$ for predicted and $s_t$ for ground-truth states.

\subsection{Two-stage Training}
A growing line of work~\citep{yu2025dynathink, zhang2026agent,
yu2025dynamind, yu2026rwml} adopts a two-stage post-training pipeline for
agentic LLMs: a \emph{world-model} stage that grounds $\pi_\theta$ in
environment dynamics, followed by a \emph{policy RL} stage that optimizes
task-success reward.

\paragraph{World-model SFT.}
The simplest instantiation finetunes $\pi_\theta$ on transition triplets to
maximize $\log \pi_\theta(s_{t+1} \mid s_{t-H:t}, a_t)$~\citep{yu2025dynathink,
zhang2026agent, yu2025dynamind, chen2025spa}. \wmsft inherits the
pitfalls of imitation learning: token-level rather than semantic
supervision, sensitivity to data quality, and model collapse / catastrophic
forgetting~\citep{yu2026rwml}.

\paragraph{World-model RL (\wmrl).}
\citet{yu2026rwml} train the world model with RL on a self-supervised \emph{sim-to-real} reward, which we call \wmrl. Given a prediction
$(\tau_t, \hat{s}_{t+1}) \sim \pi_\theta(\cdot \mid s_{t-H:t}, a_t)$, the
reward
\begin{equation}
    r^{\mathrm{WM}}
  = \mathbf{1}\!\left[\, \mathrm{sim}\!\big(E(\hat{s}_{t+1}),\, E(s_{t+1})\big) > \eta \,\right]
\end{equation}
scores embedding-space proximity to the realized next state, where
$E(\cdot)$ is a frozen embedding model, $\mathrm{sim}(\cdot,\cdot)$ a
similarity function, and $\eta$ a threshold. $\pi_\theta$ is optimized
with GRPO~\citep{shao2024deepseekmath} on its own rollouts using this
reward, thereby encouraging \emph{semantic} next-state alignment. We test $\eta \in \{ 0.5, 0.8, 0.9 \}$, and also test alternates including LLM-as-judge evaluation (no-$\eta$) and world-model transitions extracted from official ALFWorld expert data.

\paragraph{Policy RL (\prl).}
The second stage then optimizes $\pi_\theta$ end-to-end for task success with
GRPO:
\begin{multline*}
  \mathcal{L}_{\mathrm{GRPO}}
  = -\mathbb{E}_{\pi_{\theta_{\mathrm{old}}}}\!\Big[
    \min\!\big(\rho_\theta A,\, \mathrm{clip}(\rho_\theta, 1\pm\epsilon)\, A\big) \\
  - \beta\, D_{\mathrm{KL}}\!\big(\pi_\theta \,\Vert\, \pi_{\theta_{\mathrm{ref}}}\big)
  \Big],
\end{multline*}
with $\rho_\theta = \pi_\theta(y\mid x)/\pi_{\theta_{\mathrm{old}}}(y\mid x)$
and group-relative advantage $A = (r_T - \mu_G)/\sigma_G$.

\subsection{Benchmarks and Models}
\label{sec:benchmarks}

Our primary benchmark is \textbf{ALFWorld}~\citep{shridhar2021alfworld}, a text-based embodied environment
in which the agent locates and interacts with household objects to complete
natural-language tasks. It tests long-horizon planning and tracking of
action-conditioned state transitions. We use \textbf{$\tau^2$-Bench}~\citep{barres2026tau2} as a secondary test, which is an interleaved tool-use environment in which the agent alternates between tool calls and dialogue with a simulated user. 

\paragraph{Model preparation.}
We follow the training setup of~\citet{yu2026rwml}: Qwen2.5-7B-Instruct
\citep{qwen2.5} for ALFWorld (Qwen3-8B~\citep{qwen3} for
$\tau^2$-Bench). The checkpoints studied throughout the paper are re-trained in this setup: the base model, \wmrl-only, \prl-only, and sequential $\wmrl\!\to\!\prl$. Full data, hyperparameter, and prompting
details are in Appendix~\ref{app:alfworld} and ~\ref{app:tau2-replicate}.

\paragraph{Replicated performance.}
Table~\ref{tab:replicate-alfworld} reports ALFWorld results; our numbers
track~\citet{yu2026rwml}. See Appendix~\ref{app:tau2-replicate}
for $\tau^2$-Bench replication.
\begin{table}[h]
\centering

\setlength{\tabcolsep}{4pt}
\caption{Replicated success rates on ALFWorld (\%), mean $\pm$ std over 3
seeds. Our numbers closely match~\citet{yu2026rwml}.}
\label{tab:replicate-alfworld}
\begin{tabular}{lccc}
\toprule
Model & ID & OOD & AVG \\
\midrule
ReAct (base)   & 19.8$\pm$4.5 & 15.6$\pm$1.3 & 17.7$\pm$2.9 \\
\midrule
\wmsft          & 5.0$\pm$1.0  & 2.1$\pm$1.5  & 3.5$\pm$1.2  \\
\wmrl           & 33.6$\pm$3.4 & 31.8$\pm$5.3 & 32.7$\pm$3.3 \\
\midrule
\prl            & 84.6$\pm$1.3 & 71.9$\pm$2.6 & 78.3$\pm$0.9 \\
$\wmsft\!\to\!\prl$  & 66.4$\pm$1.3 & 55.2$\pm$2.0 & 60.8$\pm$1.5 \\
$\wmrl\!\to\!\prl$   & 87.0$\pm$0.4 & 81.8$\pm$0.7 & 84.4$\pm$0.3 \\
\bottomrule
\end{tabular}
\end{table}

\section{Methodology}
\label{sec:method}

We first define several measures we use to study the checkpoints mentioned.

\paragraph{$\Delta$-decomposition.}
Let $\theta_0$ denote the base-model parameters. Following the task-vector formulation of \citet{ilharco2023editing}, for each post-training
paradigm $X \in \{\wmsft,\, \wmrl,\, \prl,\, \wmrl\!\to\!\prl\}$ we obtain
trained parameters $\theta_X$ and define the additive update
$\Delta_X = \theta_X - \theta_0$. This yields four primary deltas:
$\dwmsft$, $\dwmrl$, $\dprl$ (\prl trained directly from base), and
$\dwmprl$ (the cumulative update after sequential $\wmrl\!\to\!\prl$). To
isolate the policy stage's post-\wmrl (conditional) contribution, we further define
\begin{equation}
    \ddprl = \dwmprl - \dwmrl
\end{equation}

so that $(\dwmrl,\, \ddprl)$ cleanly decomposes the sequential pipeline.
All deltas are computed module-wise on the seven trainable linear
projections of every transformer block
($\mathtt{\{q,k,v,o,up,gate,down\}\_proj}$). Since all checkpoints in our setting share the same architecture and pretrained initialization, their updates also share a common coordinate system. This facilitates a more principled weight-space comparison than for independently trained networks where parameter similarity can be distinct from functional similarity. Additionally, prior work has shown that for models fine-tuned from a common pretrained base, linear operations on weights can correspond predictably to linear changes in intermediate features~\citep{zhou2024on}, which offers further empirical support for treating their parameter updates as comparable objects.

\paragraph{Spectral interventions.}
For each module $\Delta \in \mathbb{R}^{m \times n}$ we compute the SVD
$\Delta = U \Sigma V^\top$ and then compute derived quantities defined below.

\noindent \textit{Effective rank}~\citep{roy2007effective}, how broadly $\Delta$ spreads across singular directions:
\begin{equation*}
\resizebox{1.0\linewidth}{!}{$
    \erank(\Delta)
  = \exp\!\Big(\!-\!\sum\nolimits_i p_i \log p_i\Big), \
  p_i = \sigma_i \big/ \sum\nolimits_j \sigma_j,
$}
\end{equation*}
where $\sigma_i$ are singular values of $\Delta$ and $p_i$ is the spectrum-normalized mass.

\textit{Truncated rank-$k$ approximation.} The Eckart--Young best
rank-$k$ approximation~\citep{eckart1936}
$\Delta_k = U_{:,1:k}\,\Sigma_{1:k,1:k}\,V_{:,1:k}^\top$
gives a coordinate-free notion of how much of an update's information lives
in its top-$k$ singular components.

\smallskip
\textit{Subspace projection.} For two updates $\Delta_A$, $\Delta_B$ at the
same module, let $U_A^{(k)}$ and $V_A^{(k)}$ denote the leading $k$ left
and right singular vectors of $\Delta_A$. Taking the left projection as
an example, we construct three projected
variants of $\Delta_B$: parallel, $U_A^{(k)}U_A^{(k)\top}\Delta_B$; orthogonal,
$(I-U_A^{(k)}U_A^{(k)\top})\Delta_B$; and random,
$R^{(k)}R^{(k)\top}\Delta_B$, where $R^{(k)}$ is a random orthonormal
$k$-frame of matching dimension.

The reconstructed checkpoint
$\theta_0 + \dwmrl + \mathrm{proj}(\ddprl)$ (or $+ \mathrm{proj}(\dprl)$ in the independent control) is then
re-evaluated end-to-end. 
\smallskip

\textit{Module-wise subspace overlap.}
For two leading-$k$ subspaces with orthonormal bases $B_A, B_B$, the mean
squared principal cosine $\frac{1}{k}\sum_i \sigma_i^2(B_A^\top B_B)$
measures alignment (random baseline $\approx
k/d$). This is a standard principal-angle-based subspace alignment measure~\citep{bjorck1973}.

\paragraph{Magnitude pruning of $\Delta$.}
We also retain only the top-$P\%$ of $\Delta$ entries by absolute magnitude
(zeroing the rest), following task-vector merging
conventions~\citep{yadav2023ties, yu2024dare}. This complements the
spectral analysis by operating in coordinate rather than singular-vector
space.

\paragraph{Behavioral readouts.}
We complement the geometric analyses with per-token entropy of
the model's distribution over the \emph{ground-truth} next-state
tokens on a held-out \wmrl evaluation set (lower = better
world-model calibration), and pairwise success-set overlap on
the downstream task to flag behavioral complementarity.

\section{What makes a good world model? Spectral signatures of effective two-stage training (RQ1)}
\label{sec:spectral}

\begin{figure}[h]
    \centering
    \includegraphics[width=\linewidth]{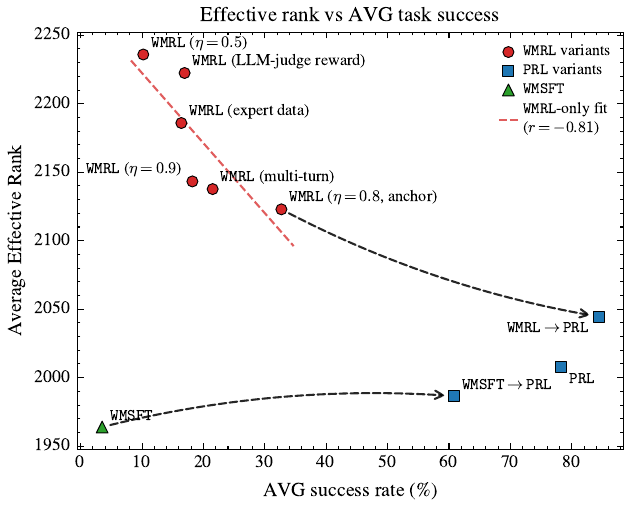}
    \caption{Effective rank vs.\ AVG success across WM-stage
    checkpoints: a $\wmsft$ baseline and a \wmrl sweep over reward
    threshold, judge type, data source, and prediction format
    (Appendix~\ref{app:wmrl-sweep}). Within \wmrl, lower effective rank
    tracks higher performance; $\wmsft$ collapses despite low rank.}
    \label{fig:er-vs-perform}
\end{figure}

\noindent Figure~\ref{fig:er-vs-perform} plots effective rank against success across WM-stage checkpoints, observing that \textbf{within RL regime, the strongest world model is low-rank.} However, \textbf{\wmsft operates in a different regime.}

\wmsft achieves lowest $\erank$ of any checkpoint yet performs catastrophically. $\wmsft\to\prl$ trails $\wmrl\to\prl$ by ${>}20$ AVG points. Thus, the two regimes are not comparable spectrally; one possible explanation is that \wmsft concentrates updates on surface textual patterns, vs. \wmrl looking for embedding-level semantic match and learning dynamics-relevant directions. We defer testing this hypothesis systematically to future work, but note that this split echoes evidence that effective fine-tuning updates concentrate in low-dimensional subspaces~\citep{aghajanyan2021intrinsic, hu2022lora} and that RL post-training has a characteristic spectral structure distinct from SFT~\citep{zhu2025path}.

\wmrl \textbf{initialization broadens the directions policy RL explores.}
Comparing $\dprl$ against $\dwmprl$, the \wmrl-initialized pipeline shows
substantially higher per-layer effective rank at every depth, with the
largest gap in the early-to-mid layers. We posit that the world-model stage opens up
directions in weight space that the policy stage subsequently
exploits---a geometric counterpart to the behavioral claim that \wmrl
contributes complementary structure rather than redundant capacity.

\paragraph{The world-model update is functionally low-rank.}
The low effective rank of strong $\wmrl$ checkpoints has a functional counterpart: truncating $\dwmrl$ to its top singular components while leaving $\ddprl$ intact incurs little performance cost. Even a rank-1 truncation (a single outer product) recovers within $\approx$10 points of the full sequential pipeline, whereas the same truncation applied to $\ddprl$ or to $\dprl$ collapses performance to near the $\wmrl$-only model at small $k$. The world-model stage's downstream value is therefore concentrated in a handful of directions, while the policy update spreads its task value more diffusely. We refer the reader to Appendix~\ref{sec:trunc} for the full results.

\section{Weight-space Relationship between $\dwmrl$ and the Policy Stage (RQ2)}
\label{sec:relationship}

Since \wmrl and \prl leave deltas of
different rank, we ask whether the directions populated by the two
stages overlap or are disjoint. Is their relationship more one of \emph{complementarity} or of
\emph{redundancy}? We identify two accounts that could explain the observed two-stage $\wmrl\!\to\!\prl$ performance gain:

\begin{figure*}[t]
  \centering
  \includegraphics[width=0.98\textwidth]{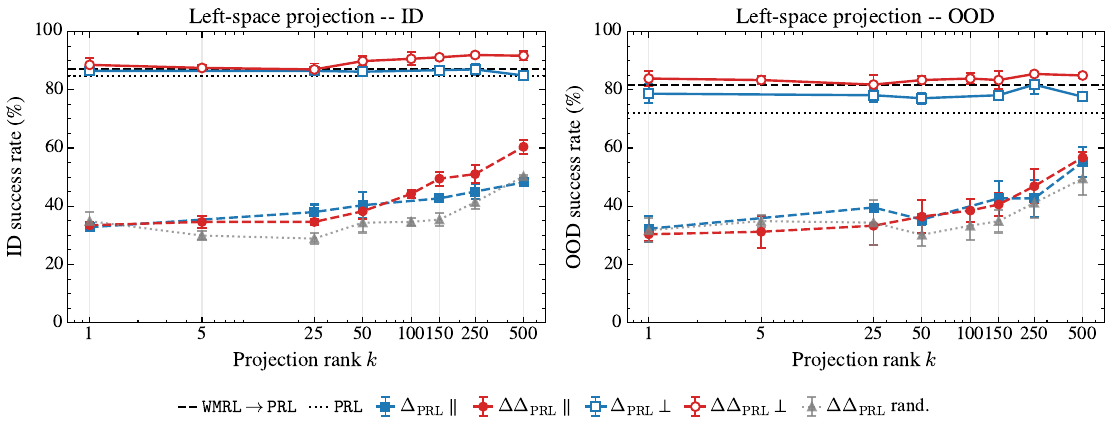}

  \caption{\textbf{Left-space (output-direction) projection on
  ALFWorld, reported separately on ID and OOD.}
  The two panels share a legend and overlay both projector variants
  on each policy update: parallel (dashed, filled marker) keeps the
  $\dwmrl$-aligned component; orthogonal (solid, hollow marker)
  keeps its complement. Blue series correspond to $\dprl$, red to
  $\ddprl$; gray is the random-subspace control for $\ddprl$.
  Black dashed and dotted lines denote $\wmrl\!\to\!\prl$ and
  \prl baselines, respectively, with values taken from the same
  ID or OOD split as the panel.}
  \label{fig:left-projection}
\end{figure*}

\begin{hypobox}{h1bg}{h1frame}
\textbf{Account A (shared / parallel).}
The policy stage refines, compresses, or warm-starts from
directions $\wmrl$ has already identified and thus its update sits largely within $\dwmrl$'s subspace. If true, keeping only the $\dwmrl$-aligned component of $\ddprl$ should recover most of the gain.

\end{hypobox}

\begin{hypobox}{h3bg}{h3frame}
\textbf{Account B (orthogonal / complementary).}
The policy stage complements \wmrl, learning an update in \emph{different} directions of weight space. If true, most of $\ddprl$'s task value should lie in the orthogonal complement of $\dwmrl$'s leading directions.
\end{hypobox}
\smallskip

We test both accounts through left/right subspace projections of $\ddprl$ and $\dprl$
(\S\ref{sec:proj-left}--\ref{sec:proj-right}), and module-wise principal-angle overlap (\S\ref{sec:overlap}).

\subsection{Left-space (Output-direction) Projection}
\label{sec:proj-left}

We compute the SVD $\dwmrl = U\Sigma V^\top$ and form
$U_k = U_{:,1:k}$. For each $k \in \{1,25,50,150,250,500\}$, we
replace $\ddprl$ with one of three projected variants: projection $U_kU_k^\top\ddprl$, $(I-U_kU_k^\top)\ddprl$, $R_kR_k^\top\ddprl$ (parallel, orthogonal, random control). We then re-evaluate
$\theta_0+\dwmrl+\mathrm{proj}(\ddprl)$.

\paragraph{Finding 1: Policy-stage output directions are largely orthogonal to $\dwmrl$.}
Removing $\dwmrl$'s top-$k$ left-singular directions from $\ddprl$
has essentially no cost on either split
(Fig.~\ref{fig:left-projection}, red hollow series): at $k{=}500$,
ID $91.7$ and OOD $84.9$, both at or slightly above the
un-projected $\wmrl\!\to\!\prl$ baseline (ID $87.0$, OOD $81.8$).
The bulk of $\ddprl$'s task-relevant signal therefore lives outside
$\dwmrl$'s principal \emph{output} subspace.

\paragraph{Finding 2: $\dwmrl$'s $U$ directions carry only marginal policy-relevant signal.}
Keeping only $\dwmrl$'s top-$k$ left-singular directions in
$\ddprl$ (red dashed series) recovers performance gradually,
reaching ID $60.4$ / OOD $56.8$ at $k{=}500$. This trails the
orthogonal projection by ${\sim}30$ points on \emph{both} splits;
a matched random projection (gray) reaches ID $50.3$ / OOD $49.5$,
only ${\sim}8$ points lower. The output directions \wmrl has
populated are only mildly enriched in policy-relevant
information.

\paragraph{Finding 3: $U$-disjointness is likely intrinsic to policy training, not only two-stage.}
The pattern holds when we project $\dprl$ (\prl trained
directly from base, never conditioned on $\dwmrl$; blue series). Orthogonal projection preserves performance (at $k{=}500$, ID
$84.9$ / OOD $77.6$, vs.\ un-projected \prl at ID $84.6$ / OOD
$71.9$ -- in fact slightly \emph{above} \prl on OOD). Parallel
projection recovers only ID $32.8$ / OOD $32.3$ at $k{=}1$ and
ID $48.2$ / OOD $55.2$ at $k{=}500$. Because $\dprl$ is computed
independently of $\dwmrl$, the orthogonality cannot be a
consequence of sequential training; it likely reflects an intrinsic difference between the preferred \emph{output} subspaces of policy and world-model learning. Per-checkpoint numbers underlying Fig.~\ref{fig:left-projection} are reported in Appendix~\ref{app:perftables} (Tabs.~\ref{tab:perf-left-dp} and \ref{tab:perf-left-ddp}).

\subsection{Right-space (Input-feature) Projection}
\label{sec:proj-right}

\S\ref{sec:proj-left} only examines \emph{output} alignment. Two updates can be orthogonal as matrices and still touch the same input features, if the rotation between their right bases happens to land on different columns of $U$. We therefore repeat the analysis on the right side, projecting onto $\dwmrl$'s leading \emph{right} singular vectors $\ddprl\, V_k V_k^\top$ and $\ddprl(I - V_k V_k^\top)$.

\begin{figure*}[t]
  \centering
  \includegraphics[width=0.98\textwidth]{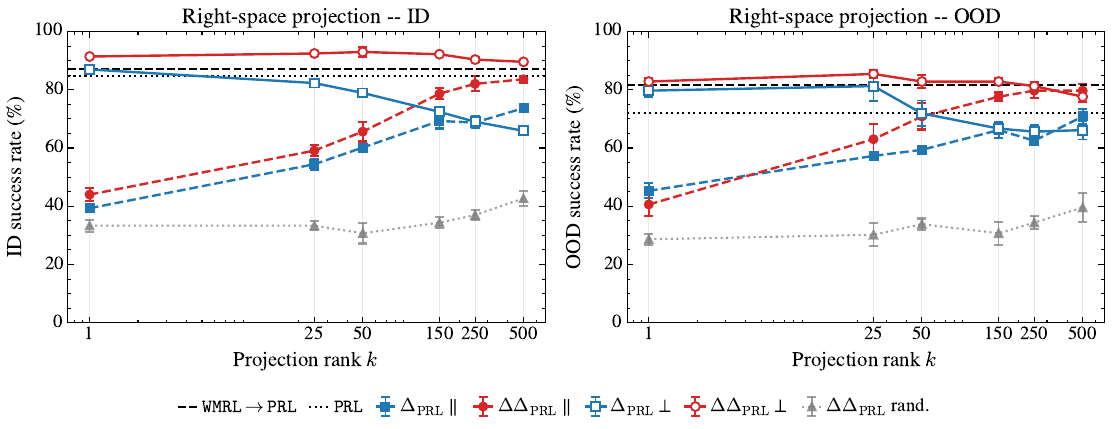}

  \caption{\textbf{Right-space (input-feature) projection on
  ALFWorld, reported separately on ID and OOD.}
  Same plotting conventions as
  Fig.~\ref{fig:left-projection}: the two panels share a legend;
  parallel projection is shown with dashed filled markers, and
  orthogonal projection with solid hollow markers. Blue series
  correspond to $\dprl$, red to $\ddprl$, and gray to the
  random-subspace control. Black dashed and dotted lines denote
  $\wmrl\!\to\!\prl$ and \prl baselines, respectively. The picture
  is mirrored relative to the left: orthogonal removal hurts
  $\dprl$ monotonically with $k$, while parallel projection of
  $\ddprl$ recovers most of the $\wmrl\!\to\!\prl$ baseline by
  $k{=}500$ on both splits.}
  \label{fig:right-projection}
\end{figure*}

\paragraph{Finding 4: In contrast, the two stages \textit{share} the right/$V$ (input-feature) subspace.}
\textit{Parallel} projection on the right
(Fig.~\ref{fig:right-projection}) is dramatically more effective than on the left: $\ddprl$ projected onto $\dwmrl$'s top right
subspace gives ID $44.0$ / OOD $40.6$ at $k{=}1$ and climbs to ID $83.6$ / OOD $79.7$ at $k{=}500$ -- within a few points of the un-projected $\wmrl\!\to\!\prl$ baseline (ID $87.0$ / OOD $81.8$) on both splits. The same operation on the left only reached ID $60.4$ / OOD $56.8$ at $k{=}500$. The random control stays near the \wmrl baseline, ruling out a generic rank-$k$ effect. This suggests \textbf{$V_{\dwmrl}$ encodes task-relevant information}.

\paragraph{Finding 5: Removing $V$ from $\dprl$ hurts performance, but not for $\ddprl$, suggesting $\ddprl$ learns robust $V$-redundant pathways.}
For $\ddprl$, removing $\dwmrl$'s leading right singular directions barely hurts: ID $91.4 \to 89.6$ (nearly flat) and OOD $82.8 \to 77.6$ (mild drop) from $k{=}1$ to $k{=}500$. Taken with Finding 4, this suggests that $\ddprl$'s task value is redundant across $\dwmrl$'s leading right directions and their complement rather than confined to either: removing one side appears to leave a functioning input pathway on the other. However, on $\dprl$, removing the same directions makes performance drop from ID $87.0$ / OOD $79.7$ at $k{=}1$ to ID $65.9$ / OOD $66.2$ at $k{=}500$, well below the un-projected \prl baseline (ID $84.6$ / OOD $71.9$) on both splits. Because $\dprl$ is trained directly from base with no exposure to $\dwmrl$, we infer that \prl-from-base independently discovers a subspace that overlaps with $\dwmrl$'s, to the point that projecting $\dwmrl$'s right directions out of $\dprl$ removes content the policy update needs to function. This adds evidence for Account A on the right side. Per-checkpoint numbers underlying
Fig.~\ref{fig:right-projection} are in
Appendix~\ref{app:perftables} (Tabs.~\ref{tab:perf-right-dp} and
\ref{tab:perf-right-ddp}).

\subsection{A Unified View of Two-stage Agentic Post-Training Geometry}
\label{sec:synth}
Together, these results point toward a coherent picture of how the two stages interact: $\dwmrl$ concentrates its update in a low-rank right subspace $V$ that also contains the policy stage's task-relevant input signal (Finding 4), policy training (even independently) writes outputs in $U$-directions disjoint from $\wmrl$ (Findings 1--3), and \textbf{sequential training composes the two components without geometric interference} (Finding 5). In other words, the geometrically identifiable value of sequential training is primarily $V$-redundancy in the downstream policy $\Delta$. Accounts A and B thus hold \textit{in tandem} on different sides. Though this establishes their weight-space roles, it does not explain how these differences manifest at deployment, or what complementarity accomplishes behaviorally. We next seek behavioral evidence to fill out this picture.

\subsection{Module-wise Subspace Overlap}
\label{sec:overlap}
\begin{figure}[h]
  \centering
  \includegraphics[width=\linewidth]{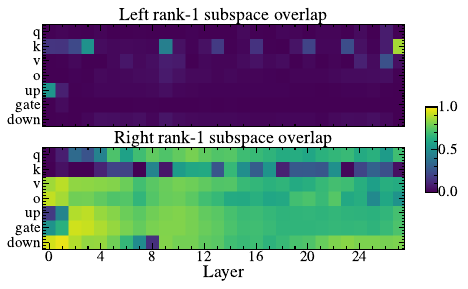}

  \caption{\textbf{Module-wise rank-$1$ subspace overlap} between
  $\dprl$ and $\ddprl$ on the left (output) vs.\ right (input-feature)
  side. }
  \label{fig:overlap}
\end{figure}
We show in Fig.~\ref{fig:overlap} that the global picture also holds at the module level. For every pair of corresponding weight matrices in $\dprl$ and
$\ddprl$, we compute the principal-angle overlap
$\bar{\sigma}^2 = \frac{1}{k}\sum_i \sigma_i^2(B_A^\top B_B)$
between their leading-$k$ left (or right) subspaces. The top panel shows the left-side overlap ($k{=}1$, mean $0.04$), nearly orthogonal, while the bottom panel shows the right-side overlap ($k{=}1$, mean $0.63$), highly aligned at dominant input direction.

\subsection{Cross-Benchmark Replication}
On $\tau^2$-Bench (with Qwen3-8B), the $U$-disjoint and $V$-shared geometric pattern replicates. Rank-1 left-subspace overlap remains near-zero, whereas the right-subspace overlap appears even more strongly than on ALFWorld. However, this is associated more modestly with end-to-end projection performance. The right-parallel projection recovers 71.8\% of oracle improvement on ALFWorld vs. 46.9\% on $\tau^2$-Bench, and we thus interpret the geometric replication as stronger than the behavioral evidence (full results in Appendix~\ref{app:tau2-block}).

\section{How does \wmrl help in practice? A rollout-time behavioral account (RQ1)}
\label{sec:how-helps}

The geometric findings raise a behavioral question: what does the \wmrl-derived structure let the agent do that \prl alone cannot? We probe this through success-set overlap and rollout-time statistics.

\paragraph{Success-set overlap.}

\begin{figure}[!htb]
    \centering
    \includegraphics[width=\linewidth]{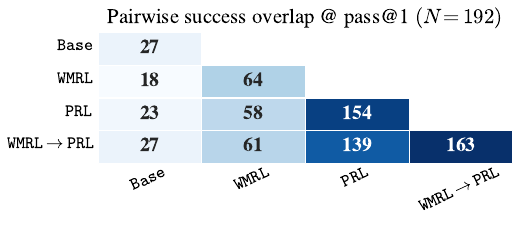}
    \caption{Success-set overlap at pass@1 on ALFWorld. Diagonal = \#samples a model passes; off-diagonal = \#samples both pass.}
    \label{fig:success-overlap}
\end{figure}

For each pair of checkpoints we compare tasks solved at pass@1 (Fig.~\ref{fig:success-overlap}). The composite $\wmrl\!\to\!\prl$ model does \emph{not} subsume either $\wmrl$'s or $\prl$'s success set: each solves a non-trivial number of tasks $\wmrl\!\to\!\prl$ fails on, while $\wmrl\!\to\!\prl$ itself picks up new successes that neither base achieves. This non-nesting rules out the simplest reading, i.e. that $\wmrl\!\to\!\prl$ is simply a refined version of $\prl$, since refinement would predict success-set containment. This suggests that what the second stage is doing on top of $\wmrl$ is not a near-copy of what $\prl$ does on its own and instead points to synergistic effects.

\paragraph{Exploration statistics.}
Table~\ref{tab:explore-stats} reports per-task averages of unique
valid actions, unique observation states, and total turns across
rollouts, with each run capped at 30 turns. Uniqueness is computed
by exact string match, and an action is valid only if it changes the
environment state. We also report the three-run union as a
cross-check.

\begin{table}[h]
\centering
\small
\setlength{\tabcolsep}{3pt}
\renewcommand{\arraystretch}{1.2}
\caption{Exploration statistics on ALFWorld (ID+OOD, $N{=}192$).
We report per-task averages of unique valid actions, unique states,
and total turns.}
\label{tab:explore-stats}
\begin{tabular}{lccc@{\hskip 0.7em}ccc}
\toprule
& \multicolumn{3}{c}{\textit{Pass@1}}
& \multicolumn{3}{c}{\textit{Pass@3 Union}} \\
\cmidrule(lr){2-4}\cmidrule(lr){5-7}
Model & $N_\text{act}$ & $N_\text{state}$ & Turns
      & $N_\text{act}$ & $N_\text{state}$ & Turns \\
\midrule
Base
& 10.8 & 14.8 & 28.0
& 18.3 & 26.7 & 82.0 \\
\wmrl
& 7.8 & 13.5 & 24.9
& 13.3 & 26.5 & 74.6 \\
\prl
& 7.1 & 8.1 & 12.6
& 8.8 & 10.3 & 37.7 \\
$\wmrl\!\to\!\prl$
& 8.4 & 9.7 & 12.4
& 11.6 & 13.9 & 37.5 \\
\bottomrule
\end{tabular}
\end{table}

\wmrl does not improve unique-action coverage over the base model, but maintains broader \emph{state} exploration. \prl, in contrast, sharply contracts both action and state coverage in exchange for shorter, more focused trajectories. The composite $\wmrl\!\to\!\prl$ inherits \prl's efficiency (similar turn count) while recovering some of \wmrl's state-coverage advantage, suggesting that the
exploration benefit of \wmrl is preserved \emph{behaviorally} through the policy stage. This rules out $\wmrl\!\to\!\prl$ being simply a more efficient $\wmrl$ or a more reward-optimized $\prl$ since each of the baselines lacks one of the two behavioral axes the composite has.

\paragraph{Next-state prediction entropy on held-out WM set.}
\begin{figure}[!htb]
    \centering
    \includegraphics[width=\linewidth]{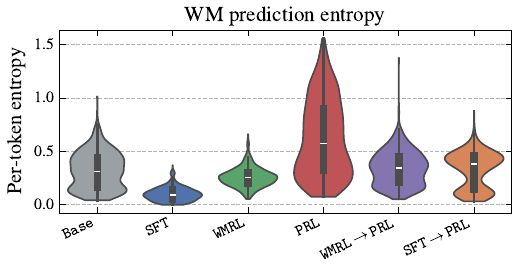}
    \caption{Per-token entropy of each checkpoint's distribution
    over the \emph{ground-truth} next-state tokens, evaluated on
    the \wmrl held-out test set ($N{=}512$ single-turn transitions) and
    averaged per sample.}
    \label{fig:entropy}
\end{figure}

Fig.~\ref{fig:entropy} reports per-token entropy on a held-out \wmrl evaluation set, scored against the \emph{ground-truth} next-state tokens. The prompt is slightly revised to directly predict the next state without thinking for fair comparison, same as $\wmsft$. Lower indicates the models are better calibrated as a world model. We observe (i) $\wmrl$ reduces next-state entropy relative to base, i.e. $\wmrl$ training makes the model a more calibrated next-state predictor, consistent with its training objective (ii) $\prl$-from-base \emph{degrades} next-state calibration, by itself actively pushing the model \textit{away} from accurate next-state prediction, and (iii) $\wmrl\!\to\!\prl$ preserves most of \wmrl's improvement. Here, the second-stage policy update does \textit{not} erase the calibration that the first-stage $\wmrl$ update put in place.

\paragraph{Summary}
Taken together, the three behavioral findings point in the same direction: $\wmrl\!\to\!\prl$ is an effective combination of two qualitatively different stages. The non-nested success sets show this at the \textit{task} level (each baseline solves tasks the composite misses, and vice versa). The exploration statistics show it at the \textit{trajectory} level (the composite combines \wmrl's state coverage with $\prl$'s efficiency). Finally, the entropy results show it at the \textit{representational} level (the composite retains the WM calibration that $\prl$-from-base implicitly degrades). We hypothesize that these three patterns are \textit{plausible} (observational) behavioral correlates of the \S\ref{sec:relationship} geometry, in which the two stages write to disjoint output subspaces (Findings 1--3) while sharing input subspaces (Finding 4), although we cannot directly establish a causal relationship. This connection is most direct for entropy preservation, since orthogonal weight updates do not overwrite each other's columns by construction. Though these properties are not uniquely predicted by the geometric patterns, they point toward future causal investigations to better characterize such effects.

\section{Preserving the World Model (RQ3)}
\label{sec:preserve}

The asymmetric geometry of \S\ref{sec:relationship} predicts that gentle preservation of $\dwmrl$ during policy training could help performance, and that extra
world-model signal should compose constructively with the policy update. We test this assumption with two interventions, one training-free and one training-time (for full details, refer to 
Appendix~\ref{app:preserve-details}):

\paragraph{Training-free: KnOTS / TIES merging of
$\mdl{\wmrl}$ and the policy model.}
We additively merge $\mdl{\wmrl}$ with either $\mdl{\prl}$ (\prl
from base) or $\mdl{\wmprl}$ (sequential $\wmrl\!\to\!\prl$) via
TIES~\citep{yadav2023ties} and KnOTS~\citep{stoica2025model}.
KnOTS factors out a singular basis shared across tasks before
applying the TIES mask; since our geometry has $V$ shared and $U$
disjoint, we test both the default $U$-sharing and a
$V$-sharing variant.

\paragraph{Training-based: online WM-SFT during \prl.}
During the second stage we additionally subsample 32 realized transitions from each step's rollouts to compute a next-state SFT loss on top of
the policy gradient, mirroring the ECHO recipe of concurrent work by \citet{shrivastava2026echo}, which learns a world model ``for free'' inside terminal-agent RL.

\begin{table}[!htb]
\centering
\small
\setlength{\tabcolsep}{4pt}
\renewcommand{\arraystretch}{1.1}
\caption{Preservation interventions on ALFWorld. For each base
pair (\prl{} or $\wmrl\!\to\!\prl$ as the second task) we list
the source checkpoints, plain TIES merging, and KnOTS-TIES in both the default $U$-sharing and our $V$-sharing variant.
The bottom row is the training-based online $\wmsft$ auxiliary loss
during \prl. \textbf{Bold}/\underline{underlined} indicates the best/baseline results within each sub-block.}
\label{tab:merge-vs-projection}
\begin{tabular}{lccc}
\toprule
Method & ID & OOD & AVG \\
\midrule
base                                          & 19.8 $\pm$ 4.5 & 15.6 $\pm$ 1.3 & 17.7 $\pm$ 2.9 \\
$\wmrl$                               & 33.6 $\pm$ 3.4 & 31.8 $\pm$ 5.3 & 32.7 $\pm$ 3.3 \\
\midrule
$\prl$                                 & \underline{84.6 $\pm$ 1.3} & \underline{71.9 $\pm$ 2.6} & \underline{78.3 $\pm$ 0.9} \\
\multicolumn{4}{l}{\textit{training free}} \\
TIES & 85.0 $\pm$ 1.7 & 72.6 $\pm$ 1.2 & 78.8 $\pm$ 0.5 \\
KnOTS (U)     & \textbf{88.0 $\pm$ 0.7} & 74.5 $\pm$ 3.2 & 81.2 $\pm$ 1.7 \\
KnOTS(V)  & 87.8 $\pm$ 2.2 & \textbf{78.7 $\pm$ 1.9} & \textbf{83.2 $\pm$ 1.9} \\
\midrule
$\wmrl\!\to\!\prl$                            & \underline{87.0 $\pm$ 0.4} & \underline{81.8 $\pm$ 0.7} & \underline{84.4 $\pm$ 0.3} \\
\multicolumn{4}{l}{\textit{training free}} \\
TIES & 87.5 $\pm$ 1.3 & 82.0 $\pm$ 0.4 & 84.8 $\pm$ 1.0\\
KnOTS(U)   & 89.8 $\pm$ 1.1 & \textbf{82.8 $\pm$ 1.3} & \textbf{86.3 $\pm$ 0.3}\\
KnOTS(V)   & 90.1 $\pm$ 1.0 & 81.8 $\pm$ 1.9 & 85.9 $\pm$ 0.7 \\
\multicolumn{4}{l}{\textit{training based}} \\
Online \wmsft  & \textbf{90.4 $\pm$ 0.4}  & 79.5 $\pm$ 0.7 & 85.0 $\pm$ 0.5\\
\bottomrule
\end{tabular}
\end{table}

\paragraph{Interventions improve over the untreated $\wmrl\!\to\!\prl$
baseline.}
On $(\mdl{\wmrl},\mdl{\prl})$, KnOTS-$V$ beats KnOTS-$U$
by $+2.0$ ($+4.2$ OOD), matching the $V$-shared/$U$-disjoint prediction: factoring out the basis that the two tasks actually share ($V$) preserves signal that the $U$ basis variant loses. On $(\mdl{\wmrl},\mdl{\wmprl})$, the $V$/$U$ gap flips but is very close; because $\dwmprl = \dwmrl + \ddprl$ already incorporates $\dwmrl$, reducing the effect of the basis choice. Both merge variants still sit $\sim\!2$ points above the untreated baseline on ALFWorld. Finally, online $\wmsft$ \textit{during} $\prl$ reaches the highest ID ($90.4$) but trails on OOD ($79.5$). The two interventions therefore complement each other: a training-free merger built on the right
geometric basis ($V$) recovers most of the gain without any further optimization, and a light training-time $\wmsft$ signal further improves ID performance.

\section{Conclusion}
The composition of world models and policies is moving from an emergent property of large models to an explicit training criterion. This shift opens two questions. One is \textit{engineering}: how should the stages be combined to produce capable agents? The other is scientific: how does a single network meaningfully compose both kinds of knowledge, and what structural properties does this produce? We offer an initial pathway for investigating both. In our experiments, we find that world-model and policy training share input-feature directions and write to largely distinct output directions, and also that sequential training makes the resulting policy less dependent on the world model's leading input directions. These findings suggest ways forward for both investigating and engineering how these two stages compose.

\paragraph{Limitations.}
Our analysis is empirical and confined to two text-only agent benchmarks and two Qwen2.5/Qwen3 models. This setting is less general than more realistic web- or computer-use agents, which often require vision-language perception, grounding, and interaction. We thus view our work as a first exploration of these phenomena. However, world-model training and evaluation for multimodal agents remain comparatively immature, making text-only environments a useful controlled setting for isolating the geometry of world-model preservation. The left/right asymmetry we identify is robust across modules and depths within these settings, but its mechanistic origin remains open. One possible contributor is a difference between target distributions of world-model and policy training. However, this pattern also appears on $\tau^2$-Bench, where both of these stages contain natural-language dialogue and tool-related content. Still, our experiments cannot yet isolate the contribution of target-distribution differences.

We also restrict our analysis to additive parameter updates relative to a single base model; extending the framework to LoRA-style decomposed updates~\cite{hu2022lora} is a natural next step. Meanwhile, our interventions are module-wise but apply the same intervention to all modules. Stronger causal tests, such as cross-module analysis~\citep{wang2023interpretability}, layer-local editing~\citep{geva-etal-2021-transformer}, activation patching~\citep{meng2022locating}, or training-time constraints on the $U/V$ subspaces~\citep{zou2023transparency}, are left for future work.

Finally, our preservation experiments instantiate only a small part of the design space suggested by the geometry. We evaluate KnOTS/TIES merging and an online WM-SFT auxiliary loss, but do not extensively tune merge hyperparameters; on $\tau^2$-Bench, we replicate only the training-free merge experiments due to the cost of running online \wmsft. More scalable preservation methods and broader benchmark/model coverage remain important next steps. We hope this exploratory study motivates more systematic investigations of world-model preservation through geometry, especially in multimodal agents, and inspires scalable methods for boosting agentic post-training.

\bibliography{custom}

\appendix

\section{ALFWorld: Implementation Details and Extended Results}
\label{app:alfworld-block}
\label{app:training}

This appendix collects (i) implementation details and
hyperparameters for the ALFWorld experiments, (ii) the \wmrl
configuration sweep that produced the cluster of checkpoints in
Fig.~\ref{fig:er-vs-perform}, (iii) the module-level
effective-rank breakdown, (iv) magnitude-pruning controls
referenced in \S\ref{sec:trunc}, and (v) the full per-checkpoint
performance tables behind the projection and pruning figures in
the main text. Throughout we follow the experimental setup
of~\citet{yu2026rwml} as closely as possible: all checkpoints
are trained from publicly released base models with GRPO for
both the \wmrl and \prl stages.

\subsection{Training and Reproduction Setup}
\label{app:alfworld}

\paragraph{Base model and data.}
We start from Qwen2.5-7B-Instruct~\citep{qwen2.5}. The \wmrl training
corpus is collected by rolling out the base model on the official
ALFWorld training split ($N{=}3$ trajectories per task, temperature
$1.0$, max $30$ steps), converting trajectories into per-step transition
triplets $\langle s_{t-H:t}, a_t, s_{t+1} \rangle$, and discarding
samples with invalid actions. Following~\citet{yu2026rwml}, we fix $H=2$ for training efficiency and perform SFT with a held-out split to get a filter model, which we then use to subsample
``too-easy'' triplets retaining $\sim 30\%$ of the original. The data size is $\sim$16K. 
\paragraph{Sim-to-real reward.}
We use Qwen3-Embedding-8B~\citep{zhang2025qwen3embed} as the frozen
encoder $E(\cdot)$ with cosine-similarity threshold $\eta = 0.8$.

\paragraph{\wmrl stage.}
$2$ epochs, learning rate $1\mathrm{e}{-6}$, effective batch size $32$,
GRPO group size $8$. Each checkpoint is trained on two NVIDIA H200 GPUs for around 8 hours.
\paragraph{\prl stage.}
GRPO with terminal task-success reward, discount $\gamma = 1.0$, maximum
$15$ steps per episode during training, $300$ update steps, group size
$8$. We use the official repo of Verl-Agent~\citep{feng2025group} for the multi-turn training\footnote{\url{https://github.com/langfengq/verl-agent}}. Each checkpoint is trained on two NVIDIA H200 GPUs for around 30 hours.

\subsection{\wmrl Setting Sweep}
\label{app:wmrl-sweep}

To probe how sensitive \wmrl is to its training-time choices---and to
assemble the cluster of \wmrl configurations plotted in
Fig.~\ref{fig:er-vs-perform}---we sweep along four axes with anchor
values ($\eta=0.8$, self-rollouts, single-turn prediction, binarized
embedding score):
\begin{itemize}[leftmargin=1.4em]
  \item \textbf{Reward threshold} $\eta \in \{0.5, 0.8, 0.9\}$,
        controlling how strict the similarity criterion is.
  \item \textbf{Data source}: self-rollouts vs.\ expert demonstrations from ALFWorld official data.
  \item \textbf{Prediction format}: single-turn vs.\ three consecutive steps'
        predictions with averaged reward (multi-state predictions with multiple corresponding actions provided)
  \item \textbf{Judge category}: LLM-as-judge~\citep{zheng2023judging}
        with Qwen3-235B-A22B-Instruct vs.\ Qwen3-8B embedding score.
\end{itemize}

\begin{table}[!htb]
\centering
\small
\setlength{\tabcolsep}{4pt}
\caption{\wmrl setting sweep on ALFWorld (\%, AVG). Each row reports the
standalone \wmrl checkpoint and the downstream $\wmrl\!\to\!\prl$ pipeline.}
\label{tab:wmrl-sweep}
\begin{tabular}{lcc}
\toprule
Setting & \wmrl & $\wmrl\!\to\!\prl$ \\
\midrule
\multicolumn{3}{l}{\textit{Anchor}} \\
- &32.7$\pm$3.3 & 84.4$\pm$0.3 \\
\midrule
\multicolumn{3}{l}{\textit{Reward threshold}} \\
$\eta=0.5$                    & 10.2$\pm$3.4 & 71.9$\pm$2.0 \\
$\eta=0.9$                    & 18.2$\pm$2.6 & 76.1$\pm$0.8 \\
\midrule
\multicolumn{3}{l}{\textit{Data source}} \\
Expert demonstrations         & 16.4$\pm$3.4 & 74.0$\pm$1.3 \\
\midrule
\multicolumn{3}{l}{\textit{Prediction format}} \\
Multi-turn                    & 21.5$\pm$1.2 & 75.6$\pm$3.1 \\
\midrule
\multicolumn{3}{l}{\textit{Judge Category}}\\

LLM-as-a-Judge                & 17.2$\pm$2.8 & 78.7$\pm$1.9 \\
\bottomrule
\end{tabular}
\end{table}

\subsection{Per-layer Spectral Picture of the Pipeline}

\begin{figure}[!htb]
    \centering
    \includegraphics[width=\linewidth]{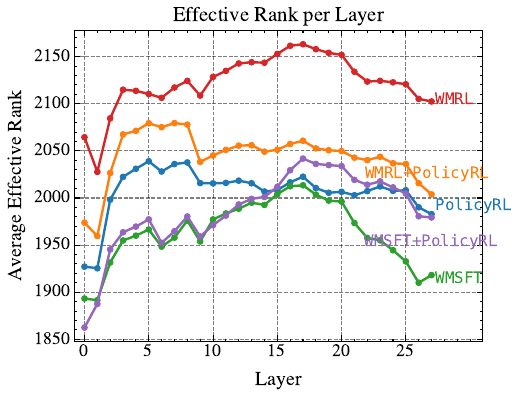}
    \caption{Per-layer effective rank of the parameter updates produced by
    each post-training paradigm, averaged across modules. The ordering of
    paradigms is preserved across depth.}
    \label{fig:erank-per-layer}
\end{figure}

At every layer, the post-training updates show a clear
spectral separation: $\dwmrl$ above $\dwmprl$ above $\dprl$, with
$\dwmsft$ far below all three (Fig.~\ref{fig:erank-per-layer}).
We use this layer-wise ordering primarily for descriptive context since many plausible accounts (e.g., shared gradient
dynamics across modules) would predict module-wise
consistency.

\subsection{Full Performance Tables for the Projection}
\label{app:perftables}

The main text reports ID and OOD success rates separately for the
projection (\S\ref{sec:relationship}) and pruning
(\S\ref{sec:trunc}, Appendix~\ref{app:topp}) experiments. For
completeness, this appendix collects the full ID, OOD, and AVG
(= Unweighted mean over ID and OOD) numbers for every checkpoint used
in those analyses, shown as the mean $\pm$
standard deviation over $3$ rollout repeats. Baselines from
Tab.~\ref{tab:replicate-alfworld} are reproduced in
Tab.~\ref{tab:perf-baselines} for convenience.

\begin{table}[!htb]
\centering
\small
\setlength{\tabcolsep}{4pt}
\caption{ALFWorld baseline reference checkpoints (reproduced from
Tab.~\ref{tab:replicate-alfworld} for cross-reference).}
\label{tab:perf-baselines}
\begin{tabular}{lccc}
\toprule
ckpt & ID & OOD & AVG \\
\midrule
ReAct(base)       & 19.8$\pm$4.5 & 15.6$\pm$1.3 & 17.7$\pm$2.9 \\
\midrule
\wmsft     & 5.0$\pm$1.0  & 2.1$\pm$1.5  & 3.5$\pm$1.2  \\
\wmrl      & 33.6$\pm$3.4 & 31.8$\pm$5.3 & 32.7$\pm$3.3 \\
\midrule
\prl       & 84.6$\pm$1.3 & 71.9$\pm$2.6 & 78.3$\pm$0.9 \\
\wmsftprl  & 66.4$\pm$1.3 & 55.2$\pm$2.0 & 60.8$\pm$1.5 \\
\wmprl     & 87.0$\pm$0.4 & 81.8$\pm$0.7 & 84.4$\pm$0.3 \\
\bottomrule
\end{tabular}
\end{table}

\paragraph{Left-space projection (\S\ref{sec:proj-left}).}
Tab.~\ref{tab:perf-left-dp} reports $\dprl$ projected onto $\dwmrl$'s
top-$k$ left singular subspace; Tab.~\ref{tab:perf-left-ddp} reports
$\ddprl$ projected onto the same.

\begin{table*}[h]
\centering
\small
\setlength{\tabcolsep}{5pt}
\caption{Left-space projection of $\dprl$ onto $\dwmrl$.
Composition: $\theta_0 + \mathrm{proj}_L(\dprl)$.
We report performance for the parallel component
($U U^\top \dprl$) and the orthogonal component
($(I-UU^\top)\dprl$).}
\label{tab:perf-left-dp}
\begin{tabular}{lccc@{\hskip 1.2em}ccc}
\toprule
& \multicolumn{3}{c}{\emph{parallel}} 
& \multicolumn{3}{c}{\emph{orthogonal}} \\
\cmidrule(lr){2-4}\cmidrule(lr){5-7}
$k$ & ID & OOD & AVG & ID & OOD & AVG \\
\midrule
1   & 32.8$\pm$1.1 & 32.3$\pm$4.5 & 32.5$\pm$2.7
    & 86.5$\pm$1.3 & 78.7$\pm$3.2 & 82.5$\pm$2.2 \\
25  & 38.0$\pm$2.6 & 39.6$\pm$0.7 & 38.8$\pm$1.5
    & 86.5$\pm$0.7 & 78.1$\pm$2.2 & 82.3$\pm$0.7 \\
50  & 40.4$\pm$4.5 & 35.4$\pm$1.9 & 37.9$\pm$1.7
    & 86.2$\pm$0.4 & 77.1$\pm$1.9 & 81.6$\pm$0.8 \\
150 & 42.7$\pm$1.3 & 42.7$\pm$5.9 & 42.7$\pm$3.2
    & 86.7$\pm$0.6 & 78.1$\pm$1.3 & 82.4$\pm$1.0 \\
250 & 45.0$\pm$2.6 & 42.7$\pm$6.4 & 43.9$\pm$3.1
    & 87.0$\pm$2.0 & 81.8$\pm$3.2 & 84.4$\pm$2.6 \\
500 & 48.2$\pm$1.0 & 55.2$\pm$5.2 & 51.7$\pm$2.7
    & 84.9$\pm$0.7 & 77.6$\pm$0.7 & 81.2$\pm$0.6 \\
\bottomrule
\end{tabular}
\end{table*}

\begin{table*}[!htb]
\centering
\small
\setlength{\tabcolsep}{4pt}
\caption{Left-space projection of $\ddprl$ onto $\dwmrl$.
Composition: $\theta_0 + \dwmrl + \mathrm{proj}_L(\ddprl)$.
Columns report the parallel component ($UU^\top\ddprl$),
orthogonal component ($(I-UU^\top)\ddprl$), and random projection
($QQ^\top\ddprl$ with $Q$ random orthonormal).}
\label{tab:perf-left-ddp}
\begin{tabular}{lccc@{\hskip 0.8em}ccc@{\hskip 0.8em}ccc}
\toprule
& \multicolumn{3}{c}{\emph{parallel}}
& \multicolumn{3}{c}{\emph{orthogonal}}
& \multicolumn{3}{c}{\emph{random}} \\
\cmidrule(lr){2-4}\cmidrule(lr){5-7}\cmidrule(lr){8-10}
$k$ & ID & OOD & AVG & ID & OOD & AVG & ID & OOD & AVG \\
\midrule
1   & 33.5$\pm$1.2 & 30.4$\pm$2.3 & 32.0$\pm$0.8 
    & 88.5$\pm$2.2 & 83.8$\pm$2.7 & 86.2$\pm$1.0
    & 34.9$\pm$3.2 & 31.8$\pm$4.1 & 33.3$\pm$1.5 \\
5   & 34.6$\pm$1.9 & 31.2$\pm$5.6 & 32.9$\pm$3.5
    & 87.5$\pm$1.1 & 83.3$\pm$1.5 & 85.4$\pm$1.1
    & 29.9$\pm$1.6 & 34.9$\pm$1.9 & 32.4$\pm$1.7 \\
25  & 34.6$\pm$1.0 & 33.3$\pm$6.5 & 34.0$\pm$3.4
    & 87.0$\pm$2.0 & 81.8$\pm$3.2 & 84.4$\pm$1.7
    & 28.9$\pm$1.9 & 34.4$\pm$7.8 & 31.6$\pm$3.9 \\
50  & 38.3$\pm$2.8 & 36.5$\pm$5.8 & 37.4$\pm$4.2
    & 89.8$\pm$1.7 & 83.3$\pm$1.5 & 86.6$\pm$0.7
    & 34.4$\pm$3.4 & 30.2$\pm$3.9 & 32.3$\pm$1.0 \\
100 & 44.3$\pm$1.5 & 38.5$\pm$3.9 & 41.4$\pm$2.2
    & 90.6$\pm$2.2 & 83.8$\pm$1.9 & 87.2$\pm$2.0
    & 34.6$\pm$1.3 & 33.3$\pm$4.8 & 34.0$\pm$2.2 \\
150 & 49.5$\pm$2.4 & 40.6$\pm$3.8 & 45.0$\pm$1.2
    & 91.2$\pm$1.0 & 83.3$\pm$3.2 & 87.2$\pm$2.1
    & 35.4$\pm$2.2 & 34.9$\pm$3.9 & 35.2$\pm$2.6 \\
250 & 51.0$\pm$3.0 & 46.9$\pm$5.8 & 49.0$\pm$3.3
    & 91.9$\pm$0.7 & 85.4$\pm$0.7 & 88.7$\pm$0.6
    & 41.4$\pm$2.2 & 41.1$\pm$5.2 & 41.3$\pm$3.7 \\
500 & 60.4$\pm$2.4 & 56.8$\pm$1.9 & 58.6$\pm$1.8
    & 91.7$\pm$1.6 & 84.9$\pm$0.7 & 88.3$\pm$1.1
    & 50.3$\pm$0.4 & 49.5$\pm$5.8 & 49.9$\pm$2.7 \\
\bottomrule
\end{tabular}
\end{table*}
\paragraph{Right-space projection (\S\ref{sec:proj-right}).}
Tab.~\ref{tab:perf-right-dp} reports $\dprl$ projected onto $\dwmrl$'s
top-$k$ right singular subspace; Tab.~\ref{tab:perf-right-ddp} reports
the same for $\ddprl$.

\begin{table*}[h]
\centering
\small
\setlength{\tabcolsep}{4pt}
\caption{Right-space projection of $\dprl$ onto $\dwmrl$.
Composition: $\theta_0 + \mathrm{proj}_R(\dprl)$.
Columns report the parallel component ($\dprl V V^\top$),
orthogonal component ($\dprl(I-VV^\top)$), and random projection
($\dprl QQ^\top$ with $Q$ random orthonormal).}
\label{tab:perf-right-dp}
\begin{tabular}{lccc@{\hskip 0.8em}ccc@{\hskip 0.8em}ccc}
\toprule
& \multicolumn{3}{c}{\emph{parallel}}
& \multicolumn{3}{c}{\emph{orthogonal}}
& \multicolumn{3}{c}{\emph{random}} \\
\cmidrule(lr){2-4}\cmidrule(lr){5-7}\cmidrule(lr){8-10}
$k$ & ID & OOD & AVG & ID & OOD & AVG & ID & OOD & AVG \\
\midrule
1   & 39.3$\pm$1.3 & 45.3$\pm$2.5 & 42.3$\pm$1.8
    & 87.0$\pm$1.6 & 79.7$\pm$2.2 & 83.3$\pm$1.8
    & 31.8$\pm$0.4 & 28.6$\pm$1.9 & 30.2$\pm$1.0 \\
25  & 54.4$\pm$1.9 & 57.3$\pm$0.7 & 55.9$\pm$0.6
    & 82.3$\pm$0.7 & 81.2$\pm$5.1 & 81.8$\pm$2.6
    & 32.5$\pm$1.6 & 32.3$\pm$1.5 & 32.4$\pm$0.3 \\
50  & 60.2$\pm$1.3 & 59.4$\pm$1.3 & 59.8$\pm$1.1
    & 78.9$\pm$1.1 & 71.9$\pm$4.4 & 75.4$\pm$2.0
    & 29.4$\pm$5.1 & 29.2$\pm$2.7 & 29.3$\pm$2.5 \\
150 & 69.3$\pm$2.6 & 66.2$\pm$2.7 & 67.7$\pm$0.8
    & 72.4$\pm$1.6 & 66.7$\pm$1.9 & 69.5$\pm$1.0
    & 34.1$\pm$1.3 & 32.8$\pm$1.3 & 33.5$\pm$1.2 \\
250 & 68.8$\pm$1.3 & 62.5$\pm$0.0 & 65.6$\pm$0.6
    & 69.0$\pm$2.0 & 65.6$\pm$2.2 & 67.3$\pm$1.9
    & 37.5$\pm$1.3 & 35.9$\pm$0.0 & 36.7$\pm$0.6 \\
500 & 73.7$\pm$1.3 & 70.8$\pm$2.7 & 72.3$\pm$2.0
    & 65.9$\pm$1.6 & 66.2$\pm$3.2 & 66.0$\pm$1.4
    & 41.7$\pm$2.6 & 37.5$\pm$2.2 & 39.6$\pm$1.8 \\
\bottomrule
\end{tabular}
\end{table*}

\begin{table*}[h]
\centering
\small
\setlength{\tabcolsep}{4pt}
\caption{Right-space projection of $\ddprl$ onto $\dwmrl$.
Composition: $\theta_0 + \dwmrl + \mathrm{proj}_R(\ddprl)$.}
\label{tab:perf-right-ddp}
\begin{tabular}{lccc@{\hskip 0.8em}ccc@{\hskip 0.8em}ccc}
\toprule
& \multicolumn{3}{c}{\emph{parallel}}
& \multicolumn{3}{c}{\emph{orthogonal}}
& \multicolumn{3}{c}{\emph{random}} \\
\cmidrule(lr){2-4}\cmidrule(lr){5-7}\cmidrule(lr){8-10}
$k$ & ID & OOD & AVG & ID & OOD & AVG & ID & OOD & AVG \\
\midrule
1   & 44.0$\pm$2.2 & 40.6$\pm$3.8 & 42.3$\pm$1.6
    & 91.4$\pm$0.6 & 82.8$\pm$1.3 & 87.1$\pm$0.8
    & 33.3$\pm$2.0 & 28.6$\pm$1.9 & 31.0$\pm$1.8 \\
25  & 59.1$\pm$1.8 & 63.0$\pm$5.2 & 61.1$\pm$2.3
    & 92.5$\pm$1.0 & 85.4$\pm$1.5 & 88.9$\pm$1.2
    & 33.3$\pm$1.5 & 30.2$\pm$3.9 & 31.8$\pm$1.5 \\
50  & 65.6$\pm$3.4 & 70.8$\pm$4.5 & 68.2$\pm$1.3
    & 93.0$\pm$1.7 & 82.8$\pm$2.2 & 87.9$\pm$0.3
    & 30.7$\pm$3.5 & 33.9$\pm$1.9 & 32.3$\pm$1.9 \\
150 & 78.7$\pm$1.8 & 77.6$\pm$1.5 & 78.1$\pm$1.7
    & 92.2$\pm$0.0 & 82.8$\pm$1.3 & 87.5$\pm$0.6
    & 34.4$\pm$1.9 & 30.7$\pm$3.9 & 32.5$\pm$1.0 \\
250 & 82.0$\pm$2.5 & 79.7$\pm$2.5 & 80.9$\pm$1.3
    & 90.4$\pm$0.7 & 81.2$\pm$1.3 & 85.8$\pm$0.4
    & 37.0$\pm$1.8 & 34.4$\pm$2.2 & 35.7$\pm$1.8 \\
500 & 83.6$\pm$1.1 & 79.7$\pm$2.2 & 81.6$\pm$1.5
    & 89.6$\pm$0.7 & 77.6$\pm$1.9 & 83.6$\pm$0.6
    & 42.7$\pm$2.6 & 39.6$\pm$4.8 & 41.1$\pm$1.4 \\
\bottomrule
\end{tabular}
\end{table*}

\subsection{Robustness Check: rank-$k$ Pruning Experiments}
\label{sec:trunc}

\begin{figure}[!htb]
  \centering
  \includegraphics[width=\linewidth]{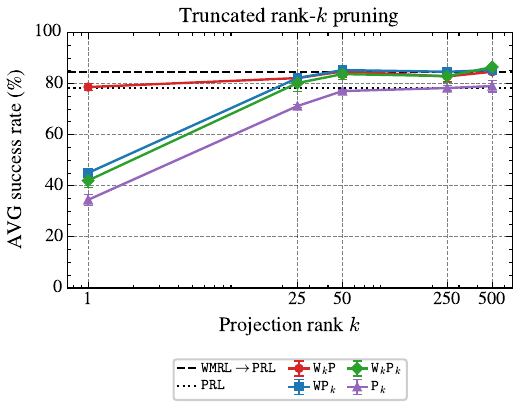}
  \caption{\textbf{Rank-$k$ truncation pruning of $\Delta$s.}
Across $k \in \{1,25,50,250,500\}$, pruning $\dwmrl$ is nearly
harmless, whereas pruning $\ddprl$ or $\dprl$ causes a sharp
small-rank drop.}
  \label{fig:prune-wkpk}
\end{figure}

The geometric picture of \S\ref{sec:relationship} -- $\dwmrl$ suggests a low-rank, input-aligned scaffold that the policy stage builds
on through output-orthogonal directions -- makes a prediction about each stage's update in isolation: $\dwmrl$
should be highly truncation-tolerant, while $\dprl$ -- a
from-base update that has to rediscover the same hypothetical scaffold --
should not be. We test these predictions through rank-$k$
truncation, following prior work on low-rank approximations of
neural network weights for compression and
adaptation~\citep{denton2014low, aghajanyan2021intrinsic, hu2022lora}
and on rank-based compression of LLM
weights~\citep{ashkboos2024slicegpt, sharma2024laser}. 

Concretely, for each $k$ and each $\Delta$ $\in$ $\{\dwmrl,\, \ddprl,\, \dprl\}$  we form the truncated rank-$k$ approximation $\Delta_k$
and evaluate four variants: $\mathtt{W}_k\mathtt{P}$, which prunes
$\dwmrl$ only; $\mathtt{W}\mathtt{P}_k$, which prunes $\ddprl$ only;
$\mathtt{W}_k\mathtt{P}_k$, which prunes both updates; and
$\mathtt{P}_k$, which directly prunes the from-base update $\dprl$. Tab.~\ref{tab:perf-wkpk} reports the per-$k$ ID, OOD, and AVG numbers
for the four rank-truncation regimes and Fig.~\ref{fig:prune-wkpk} displays the trend.

\paragraph{$\dwmrl$ is highly compressible.}
$\mathtt{W}_k\mathtt{P}$ is almost flat across ranks
(Fig.~\ref{fig:prune-wkpk}): even replacing $\dwmrl$ by its rank-$1$
truncation while keeping $\ddprl$ intact yields ID $77.6$ / OOD $79.7$,
within $\sim\!10$ points of the full $\wmrl\!\to\!\prl$ model
(ID $87.0$, OOD $81.8$). Thus, most of the downstream utility of the
WM stage is concentrated in very few directions.

\paragraph{$\ddprl$ and $\dprl$ require higher rank.}
In contrast, $\mathtt{W}\mathtt{P}_k$ drops to ID $43.8$ / OOD $46.4$
at $k{=}1$ and recovers only by moderate ranks, showing that
$\ddprl$ spreads its task value across many singular components.
$\mathtt{W}_k\mathtt{P}_k$ closely tracks $\mathtt{W}\mathtt{P}_k$,
indicating that $\ddprl$, rather than $\dwmrl$, is the bottleneck.
The direct from-base update is densest: $\mathtt{P}_k$ is worst at
every $k$, with rank-$1$ performance near the $\wmrl$-only model and rank
$500$ only matching unpruned \prl.

\paragraph{Robustness with respect to the geometric picture.}
These trends match the geometry in \S\ref{sec:relationship}:
$\dwmrl$ provides a low-rank scaffold that the policy stage can reuse,
whereas \prl from base must construct that scaffold and the task-specific
correction jointly. The similar pattern appears under magnitude pruning:
top-$p$ pruning recovers $\ddprl$ to the $\wmrl\!\to\!\prl$ baseline
best with only $10\%$ of entries retained, while pruning $\dprl$ yields best performance with
$50\%$ (Appendix~\ref{app:topp}).
\begin{table}[t]
\centering
\footnotesize
\setlength{\tabcolsep}{5pt}
\renewcommand{\arraystretch}{1.0}
\caption{Truncated rank-$k$ pruning of $\Delta$s
(\S\ref{sec:trunc}). $\mathtt{W}_k\mathtt{P}$:
$\theta_0 + (\dwmrl)_k + \ddprl$;
$\mathtt{W}\mathtt{P}_k$: $\theta_0 + \dwmrl + (\ddprl)_k$;
$\mathtt{W}_k\mathtt{P}_k$: $\theta_0 + (\dwmrl)_k + (\ddprl)_k$;
$\mathtt{P}_k$: $\theta_0 + (\dprl)_k$.}
\label{tab:perf-wkpk}
\begin{tabular}{lcccc}
\toprule
Method & $k$ & ID & OOD & AVG \\
\midrule
\multirow{5}{*}{$\mathtt{W}_k\mathtt{P}$}
& 1   & 77.6$\pm$1.0 & 79.7$\pm$2.2 & 78.7$\pm$1.3 \\
& 25  & 82.0$\pm$0.6 & 82.3$\pm$1.9 & 82.2$\pm$1.3 \\
& 50  & 85.7$\pm$1.6 & 83.8$\pm$3.0 & 84.8$\pm$2.2 \\
& 250 & 86.5$\pm$1.8 & 79.2$\pm$2.7 & 82.8$\pm$2.2 \\
& 500 & 86.5$\pm$1.0 & 82.8$\pm$2.2 & 84.6$\pm$0.8 \\
\midrule
\multirow{5}{*}{$\mathtt{W}\mathtt{P}_k$}
& 1   & 43.8$\pm$2.9 & 46.4$\pm$0.7 & 45.0$\pm$1.4 \\
& 25  & 85.7$\pm$2.2 & 78.7$\pm$1.5 & 82.2$\pm$0.7 \\
& 50  & 88.3$\pm$1.1 & 82.3$\pm$0.7 & 85.3$\pm$0.9 \\
& 250 & 88.0$\pm$1.0 & 81.2$\pm$2.2 & 84.6$\pm$0.8 \\
& 500 & 88.3$\pm$1.9 & 82.3$\pm$1.5 & 85.3$\pm$1.6 \\
\midrule
\multirow{5}{*}{$\mathtt{W}_k\mathtt{P}_k$}
& 1   & 40.4$\pm$2.4 & 43.8$\pm$3.4 & 42.1$\pm$2.7 \\
& 25  & 81.2$\pm$3.2 & 79.2$\pm$3.9 & 80.2$\pm$3.2 \\
& 50  & 85.7$\pm$2.9 & 81.8$\pm$1.5 & 83.7$\pm$2.1 \\
& 250 & 85.2$\pm$1.9 & 80.7$\pm$3.2 & 82.9$\pm$2.6 \\
& 500 & 88.0$\pm$1.0 & 84.9$\pm$1.9 & 86.5$\pm$0.5 \\
\midrule
\multirow{5}{*}{$\mathtt{P}_k$}
& 1   & 34.1$\pm$5.0 & 34.9$\pm$0.7 & 34.5$\pm$2.1 \\
& 25  & 77.3$\pm$1.7 & 65.1$\pm$1.5 & 71.2$\pm$0.7 \\
& 50  & 81.8$\pm$3.1 & 72.4$\pm$4.1 & 77.1$\pm$0.5 \\
& 250 & 84.6$\pm$2.2 & 71.9$\pm$1.3 & 78.3$\pm$0.9 \\
& 500 & 84.1$\pm$1.5 & 74.0$\pm$5.8 & 79.0$\pm$2.2 \\
\bottomrule
\end{tabular}
\end{table}

\subsection{Magnitude Pruning of $\Delta$s}
\label{app:topp}

\begin{figure}[h]
  \centering
  \includegraphics[width=\linewidth]{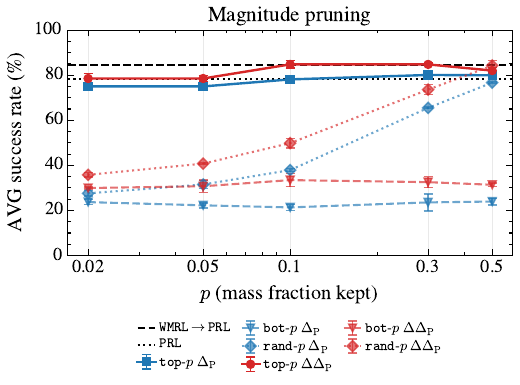}
  \caption{Magnitude pruning on $\dprl$ (blue) and $\ddprl$ (red),
  comparing the \emph{top-$p$} selection rule (solid) against
  \emph{bottom-$p$} (dashed) and \emph{random-$p$} (dotted)
  baselines. Top-$p$ keeps the entries of $\Delta$ with the largest
  absolute magnitude; bottom-$p$ keeps the smallest; random-$p$ is a
  random subset of the same mass. Top-$p$ on $\ddprl$ performs best at $p=0.1$; top-$p$ on $\dprl$ needs $p{\geq}0.3$ to reach the maxima; bottom-$p$ is near-baseline for all $p$; random-$p$ saturates only as $p \to 0.5$.}
  \label{fig:prune-topp}
\end{figure}

As a coordinate-space complement to the spectral analyses, we apply
classical magnitude pruning~\citep{han2015pruning, frankle2019lottery} --
also a central primitive in recent LLM weight pruning methods such as
SparseGPT~\citep{frantar2023sparsegpt}, Wanda~\citep{sun2024wanda} and
LLM-Pruner~\citep{ma2023llmpruner} -- to $\dprl$ and $\ddprl$,
retaining only a $p$ fraction of the $\Delta$ entries (zeroing the rest)
chosen by three different selection rules and re-evaluating task success:
(i) \emph{top-$p$}, keeping the entries with the largest absolute
magnitude; (ii) \emph{bottom-$p$}, keeping the entries with the
smallest absolute magnitude (i.e.\ the lowest-magnitude tail);
(iii) \emph{random-$p$}, keeping a random $p$ fraction. The latter
two serve as baselines that isolate, respectively, the effect of
discarding the high-magnitude head and the effect of total mass alone. Per-$p$ ID, OOD, and AVG numbers for all three selection rules are in Tab.~\ref{tab:perf-topp}. The \texttt{sign\_only} row keeps only the sign of every entry and rescales magnitudes to a fixed constant. The curves are also shown in Fig.~\ref{fig:prune-topp} to better track the trend.

Three patterns are consistent with the spectral story:
(i) top-$p$ on $\ddprl$ reaches the best performance with $p=0.1$, while the curve for $\dprl$ still goes up before $p=0.5$, indicating that the update of $\ddprl$ is more compact. This is consistent with $\ddprl$'s smaller,
output-orthogonal spectral role on top of $\dwmrl$.
(ii) Bottom-$p$ collapses both $\dprl$ and $\ddprl$ sharply for
every $p$ -- the high-magnitude entries discarded by bottom-$p$ are
precisely the ones that carry the task signal, complementing the
top-$p$ result.
(iii) Random-$p$ recovers performance roughly in proportion to mass:
$\ddprl$ random-$p$ matches top-$p$ only as $p \to 0.5$, while
$\dprl$ random-$p$ trails top-$p$ by a wide margin at small $p$. So
the gain of top-$p$ over random-$p$, at any given $p$, gives a
coordinate-level analogue of the spectral concentration story --
$\ddprl$'s top-magnitude entries dominate downstream performance much
more than $\dprl$'s.

\begin{table*}[t]
\centering
\small
\setlength{\tabcolsep}{4pt}
\caption{Magnitude pruning of $\Delta$s on ALFWorld. $p$ is the
fraction of entries retained. Left block prunes $\dprl$; right
block prunes $\ddprl$. Each section corresponds to one selection
rule, followed by the sign-only baseline.}
\label{tab:perf-topp}
\begin{tabular}{lccc@{\hskip 1.2em}ccc}
\toprule
& \multicolumn{3}{c}{$\dprl$ pruned}
& \multicolumn{3}{c}{$\ddprl$ pruned} \\
\cmidrule(lr){2-4}\cmidrule(lr){5-7}
$p$ & ID & OOD & AVG & ID & OOD & AVG \\
\midrule
\multicolumn{7}{c}{\emph{top-$p$}} \\
0.02 & 75.5$\pm$0.4 & 74.5$\pm$1.5 & 75.0$\pm$0.8
     & 82.0$\pm$5.7 & 75.0$\pm$2.2 & 78.5$\pm$2.2 \\
0.05 & 74.5$\pm$1.3 & 75.5$\pm$0.7 & 75.0$\pm$0.3
     & 82.0$\pm$3.5 & 75.0$\pm$2.5 & 78.5$\pm$1.4 \\
0.10 & 82.8$\pm$2.2 & 73.4$\pm$3.4 & 78.1$\pm$1.3
     & 88.3$\pm$1.3 & 81.2$\pm$2.2 & 84.8$\pm$1.7 \\
0.30 & 85.7$\pm$1.0 & 74.5$\pm$0.7 & 80.1$\pm$0.3
     & 88.8$\pm$0.4 & 80.7$\pm$1.5 & 84.8$\pm$0.6 \\
0.50 & 85.4$\pm$1.9 & 74.5$\pm$0.7 & 80.0$\pm$1.0
     & 87.0$\pm$1.8 & 77.1$\pm$3.9 & 82.0$\pm$1.4 \\
\midrule
\multicolumn{7}{c}{\emph{bottom-$p$}} \\
0.02 & 24.0$\pm$1.0 & 23.4$\pm$1.3 & 23.7$\pm$1.1
     & 32.3$\pm$3.2 & 27.6$\pm$3.2 & 29.9$\pm$0.4 \\
0.05 & 23.7$\pm$0.7 & 20.8$\pm$1.9 & 22.3$\pm$1.1
     & 31.8$\pm$3.0 & 29.7$\pm$4.6 & 30.7$\pm$2.7 \\
0.10 & 23.4$\pm$2.3 & 19.3$\pm$2.7 & 21.4$\pm$1.3
     & 34.1$\pm$2.7 & 32.8$\pm$3.4 & 33.5$\pm$2.9 \\
0.30 & 23.7$\pm$4.2 & 23.4$\pm$3.4 & 23.6$\pm$3.8
     & 34.9$\pm$3.7 & 30.2$\pm$3.9 & 32.5$\pm$2.4 \\
0.50 & 23.4$\pm$1.1 & 24.5$\pm$1.9 & 24.0$\pm$1.4
     & 34.6$\pm$3.0 & 28.1$\pm$2.5 & 31.4$\pm$0.7 \\
\midrule
\multicolumn{7}{c}{\emph{random-$p$}} \\
0.02 & 25.5$\pm$2.0 & 29.7$\pm$2.2 & 27.6$\pm$1.2
     & 39.3$\pm$1.5 & 32.3$\pm$2.7 & 35.8$\pm$1.0 \\
0.05 & 29.7$\pm$1.9 & 33.3$\pm$1.9 & 31.5$\pm$1.8
     & 43.0$\pm$2.2 & 38.5$\pm$1.9 & 40.8$\pm$0.4 \\
0.10 & 36.2$\pm$1.9 & 39.6$\pm$0.7 & 37.9$\pm$0.6
     & 52.9$\pm$2.9 & 46.9$\pm$3.4 & 49.9$\pm$2.1 \\
0.30 & 66.9$\pm$2.7 & 64.1$\pm$2.5 & 65.5$\pm$0.4
     & 74.2$\pm$0.6 & 72.9$\pm$3.9 & 73.6$\pm$2.3 \\
0.50 & 80.0$\pm$0.7 & 73.4$\pm$0.0 & 76.7$\pm$0.4
     & 87.2$\pm$2.0 & 81.2$\pm$2.5 & 84.2$\pm$2.2 \\
\midrule
\texttt{sign\_only}
     & 85.2$\pm$1.7 & 74.5$\pm$3.7 & 79.8$\pm$1.4
     & 83.8$\pm$1.3 & 80.2$\pm$1.9 & 82.0$\pm$1.1 \\
\bottomrule
\end{tabular}
\end{table*}

\section{$\tau^2$-Bench: Cross-task Replication}
\label{app:tau2-block}

This appendix reports a partial replication of the ALFWorld
findings on $\tau^2$-Bench with a different base model
(Qwen3-8B). \S\ref{app:tau2-replicate} describes the training
setup and end-to-end performance numbers, and
\S\ref{app:qwen3-replication} replicates the subspace
asymmetry of \S\ref{sec:relationship} on this model/task pair.

\subsection{Training and Reproduction Setup}
\label{app:tau2-replicate}

\paragraph{Base model and data.}
We start from Qwen3-8B~\citep{qwen3}. Following~\citet{yu2026rwml}, we fix the history length $H=5$ to deal with more complex scenarios. As $\tau^2$-Bench provides
only $178$ training tasks, we roll out $6$ trajectories per task --
three with GPT-4.1 as the user simulator and three with
Qwen3-235B-A22B-Instruct -- and convert trajectories into transition
triplets. To prevent the model from memorizing database contents, we
replace tool-response values with their OpenAPI schemas.

\paragraph{Sim-to-real reward.}
Following~\citet{yu2026rwml}: embedding-based binary similarity for
natural-language user responses (Qwen3-Embedding-8B, $\eta = 0.6$), and
a discretised ROUGE-L score for structured tool responses.

\paragraph{\wmrl stage.}
$2$ epochs, learning rate $1\mathrm{e}{-6}$, effective batch size $32$,
GRPO group size $16$. We use similar filtering pipeline as~\citet{yu2026rwml}. Each checkpoint is trained on two NVIDIA H200 GPUs for around 10 hours. The data size is $\sim$6K.

\paragraph{\prl stage.}
GRPO with terminal task-success reward, $\gamma = 1.0$, max $30$ steps
per episode, $200$ update steps, group size $8$. We use
Qwen3-235B-A22B-Instruct as the user simulator throughout. We also use the modified Verl-Agent GitHub repo for training and each checkpoint is trained on four NVIDIA H200 GPUs for around 100 hours. 

\begin{table}[!htb]
\centering
\small
\setlength{\tabcolsep}{4pt}
\caption{Replicated $\tau^2$-Bench performance. AVG@3 is mean $\pm$ std
success rate over 3 seeds across retail, telecom, and airline; pass@$k$
\citep{chen2021codex} is the probability that at least one of $k$
rollouts succeeds.}
\label{tab:replicate-tau2}
\begin{tabular}{lccc}
\toprule
Model & AVG@3 & pass@1 & pass@3 \\
\midrule
ReAct (base)   & $33.7\pm2.5$ & $37.0$ & $49.0$ \\
\midrule
\wmsft          & $26.0\pm4.3$ & $30.0$ & $46.0$ \\
\wmrl           & $36.3\pm0.5$ & $36.0$ & $52.0$ \\
\midrule
\prl            & $39.3\pm2.6$ & $38.0$ & $58.0$ \\
$\wmsft\!\to\!\prl$  & $38.3\pm4.5$ & $41.0$ & $59.0$ \\
$\wmrl\!\to\!\prl$   & $45.0\pm1.6$ & $43.0$ & $64.0$ \\
\bottomrule
\end{tabular}
\end{table}

\subsection{Per-layer Effective Rank}
\label{app:qwen3-erank-per-layer}

\begin{figure}[!htb]
  \centering
  \includegraphics[width=\linewidth]{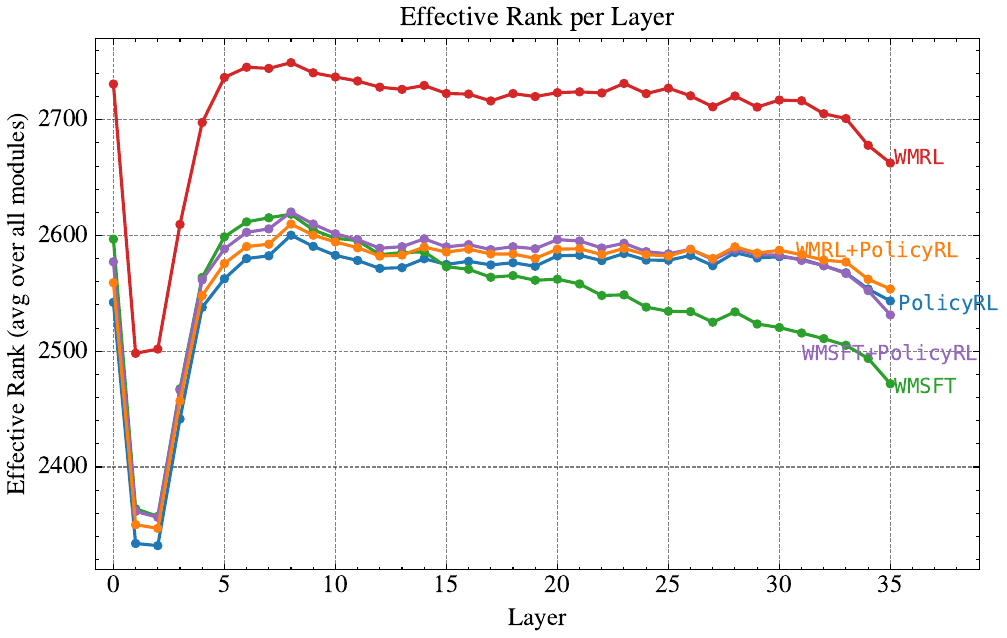}
  \caption{\textbf{Per-layer $\erank(\Delta)$ of post-training
  on Qwen3-8B + $\tau^2$-Bench}, averaged across
  all target modules. Same plot format as the main-text
  Fig.~\ref{fig:erank-per-layer}: \wmrl (red) keeps the highest
  per-layer rank; the RL-after-WM variants
  ($\wmrl\!\to\!\prl$, $\wmsft\!\to\!\prl$) and \prl cluster
  closely in the middle; \wmsft (green) is comparable in the
  early layers but drifts below the RL-trained checkpoints in the
  deeper layers.}
  \label{fig:qwen3-erank-per-layer}
\end{figure}

We reproduce the main-text per-layer effective-rank plot
(Fig.~\ref{fig:erank-per-layer}) on Qwen3-8B + $\tau^2$-Bench
checkpoints in Fig.~\ref{fig:qwen3-erank-per-layer}, using the same
plot format (one line per checkpoint, averaged across all target modules). The qualitative ordering is preserved across
all $36$ layers: \wmrl sits visibly above the RL-stage checkpoints
(\prl, $\wmrl\!\to\!\prl$, $\wmsft\!\to\!\prl$, clustered in
the middle), and \wmsft trails them in the deeper layers --
matching the Qwen2.5/ALFWorld pattern on a different model and
benchmark.

\subsection{Cross-model Replication of the Subspace Asymmetry}
\label{app:qwen3-replication}

The main-text geometric analyses in \S\ref{sec:relationship} are
conducted on Qwen2.5-7B-Instruct fine-tuned on ALFWorld. To check
that the U-disjoint / V-shared asymmetry is not an artefact of a
single model or task, we replicate the module-wise subspace overlap
measurement on a different model and benchmark: Qwen3-8B fine-tuned
on $\tau^2$-Bench. The fine-tuning recipe follows the same two-stage
pipeline ($\wmrl$ followed by $\prl$), so we can form the same
$\dprl$ and $\ddprl$ updates and compute their leading-$k$ left and
right subspace overlap module by module, exactly as in
\S\ref{sec:overlap}.

\paragraph{Headline numbers (Tab.~\ref{tab:qwen3-overlap}).}
The asymmetry holds, and is in fact slightly sharper on Qwen3.
Mean rank-$1$ \emph{left} overlap is $0.04$ on both models
(essentially identical), while mean rank-$1$ \emph{right} overlap is
$0.81$ on Qwen3 versus $0.63$ on Qwen2.5. The Frobenius cosine
between $\dprl$ and $\ddprl$ is $0.015$ on Qwen3 (vs.\ $0.028$ on
Qwen2.5), confirming that the two updates are close to orthogonal as
matrices on both models.

\begin{table}[t]
\centering
\small
\setlength{\tabcolsep}{4pt}
\renewcommand{\arraystretch}{1.1}
\caption{Cross-model replication of the module-wise subspace overlap
($\dprl$ vs.\ $\ddprl$) and Frobenius cosine, averaged over all
target modules in each model.}
\label{tab:qwen3-overlap}
\begin{tabular}{lcc}
\toprule
metric & Qwen2.5-7B & Qwen3-8B \\
& (ALFWorld) & ($\tau^2$-Bench) \\
\midrule
n\_modules                          & 196   & 252 \\
Frobenius cosine, mean              & 0.028 & 0.015 \\
\midrule
\multicolumn{3}{l}{\emph{Left subspace overlap}} \\
$k{=}1$,  mean                      & 0.036 & 0.043 \\
$k{=}50$, mean                      & 0.158 & 0.157 \\
$k{=}250$, mean               & 0.239 & 0.238 \\
\midrule
\multicolumn{3}{l}{\emph{Right subspace overlap}} \\
$k{=}1$,  mean                      & 0.635 & \textbf{0.811} \\
$k{=}50$, mean                      & 0.411 & 0.508 \\
$k{=}250$, mean               & 0.393 & 0.416 \\
\bottomrule
\end{tabular}
\end{table}

\begin{figure}[h]
  \centering
  \includegraphics[width=\linewidth]{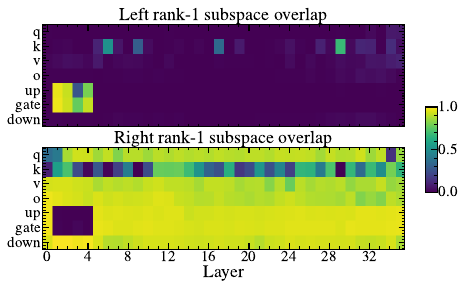}

  \caption{\textbf{Module-wise rank-$1$ subspace overlap between
  $\dprl$ and $\ddprl$ on Qwen3-8B + $\tau^2$-Bench.}
  The top panel shows the left-side overlap ($k{=}1$, mean $0.04$),
  nearly orthogonal, while the bottom panel shows the right-side
  overlap ($k{=}1$, mean $0.81$), highly aligned at the dominant
  input direction -- matching the Qwen2.5/ALFWorld pattern.}
  \label{fig:qwen3-overlap}
\end{figure}

\paragraph{Module-wise heatmaps.}
Fig.~\ref{fig:qwen3-overlap} reproduces the rank-$1$ left/right
overlap heatmaps of \S\ref{sec:overlap} on Qwen3. Visually the
pattern is the same as on Qwen2.5 (cf.\ Fig.~\ref{fig:overlap}):
the left heatmap is uniformly dark (near-zero overlap on output
directions) while the right heatmap is uniformly bright (high
overlap on input-feature directions). A small handful of early-FFN
modules (up\_proj/gate\_proj in layers 1--4) show the reverse
pattern -- shared on the left, disjoint on the right -- which we
note but do not interpret further here; outside this small block,
the U-disjoint / V-shared geometry holds module-wise across all $36$
layers.

\begin{figure}[h]
  \centering
  \includegraphics[width=\linewidth]{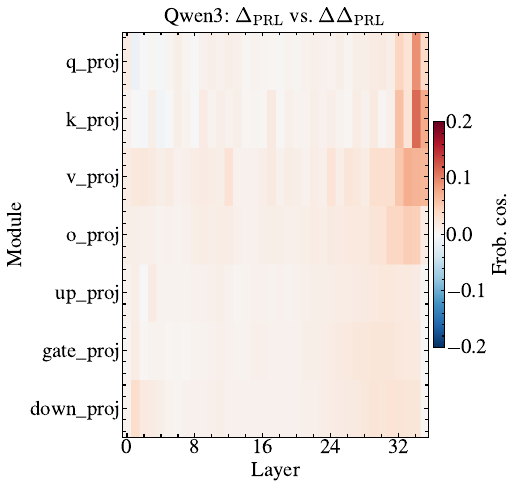}
  \caption{Per-module Frobenius cosine similarity between $\dprl$
  and $\ddprl$ on Qwen3-8B + $\tau^2$-Bench. The vast majority of
  modules are at or near zero; only a small subset of attention
  projections in the last three layers shows mild positive
  similarity ($\lesssim 0.10$). Module-wise mean is $0.015$.}
  \label{fig:qwen3-frob}
\end{figure}

\paragraph{Does the geometric asymmetry translate to a behavioral
asymmetry?} To check whether the U-disjoint / V-shared geometry
reproduces \emph{behaviorally} on Qwen3, we run the same
parallel/orthogonal projection sweep that produced the main-text
left/right curves on the Qwen3 + $\tau^2$-Bench checkpoints, and then we compare
the $k{=}1$ and $k{=}250$ rows against the Qwen2.5/ALFWorld numbers
in Tab.~\ref{tab:qwen3-perf}.

\begin{table}[t]
\centering
\small
\setlength{\tabcolsep}{4pt}
\renewcommand{\arraystretch}{1.05}
\caption{Cross-model behavioral projection results.
Numbers are end-to-end success rate (\%) averaged over $3$ repeats.
Same projection sweep as Tabs.~\ref{tab:perf-left-dp}
and~\ref{tab:perf-right-dp}, restricted to $k{=}1$ and
$k{=}250$. Geometric asymmetry replicates cleanly on Qwen3
(Tab.~\ref{tab:qwen3-overlap}), but the behavioral separation
between parallel and orthogonal projections is much smaller, for
reasons we discuss below.}
\label{tab:qwen3-perf}
\begin{tabular}{lcc}
\toprule
 & Qwen2.5-7B & Qwen3-8B \\
 & (ALFWorld) & ($\tau^2$-Bench) \\
\midrule
base                              & 17.7 & 33.7 \\
$\dwmrl$ alone                    & 32.7 & 36.3 \\
$\dprl$ alone                     & 78.3 & 39.3 \\
$\dwmrl{+}\ddprl$ (oracle)         & 84.4 & 45.0 \\ 
\midrule
\multicolumn{3}{l}{\emph{Left-space proj. of $\dprl$ onto $\dwmrl$}} \\
parallel,    $k{=}1$              & 32.5 & 34.0 \\
parallel,    $k{=}250$            & 43.9 & 35.0 \\
orthogonal,  $k{=}1$              & 82.5 & 40.3 \\
orthogonal,  $k{=}250$            & 84.4 & 40.7 \\
\midrule
\multicolumn{3}{l}{\emph{Right-space proj. of $\dprl$ onto $\dwmrl$}} \\
parallel,   $k{=}1$               & 42.3 & 35.3 \\
parallel,   $k{=}250$             & 65.6 & 39.0 \\
orthogonal, $k{=}1$               & 83.3 & 39.7 \\
orthogonal, $k{=}250$             & 67.3 & 40.7 \\
\bottomrule
\end{tabular}
\end{table}

\paragraph{Geometry replicates}
The qualitative direction is preserved: on Qwen3, orthogonal-to-$\dwmrl$
projections of $\dprl$ at $k{=}1$ are still higher than parallel
ones ($40.3$ vs.\ $34.0$, gap $+6.3$), and the parallel projection
sits at the base+$\dwmrl$ level rather than near the oracle -- the
sign of the asymmetry matches Qwen2.5. What is different is the
\emph{magnitude}: on Qwen2.5 the $k{=}1$ orth$-$par gap is
$+50$ points, on Qwen3 it is $+6$. Three factors might explain most of
this attenuation:
\begin{itemize}\itemsep0pt\topsep2pt\parsep0pt
\item \emph{Smaller dynamic range for $\dprl$.} On Qwen2.5,
  $\dprl$ buys $+60.6$ points over base ($17.7 \to 78.3$); on
  Qwen3 it buys only $+5.6$ ($33.7 \to 39.3$). The behavioral
  signal that any projection of $\dprl$ can corrupt or preserve is
  proportionally smaller.
\item \emph{Smaller two-stage gain over $\dprl$.} The
  $\dwmrl{+}\ddprl$ oracle is only $+5.7$ above $\dprl$ alone on
  Qwen3 (vs.\ $+6.1$ on Qwen2.5 in absolute, but starting from a
  far higher base) -- and as a fraction of headroom this is much
  thinner. Recovering even $80\%$ of that gain is a $\sim\!4.6$
  point swing that competes with per-repeat noise of
  $\pm 1$--$4$ points.
\item \emph{Lower absolute success rates and a harder task.}
  $\tau^2$-Bench operates in the $30$--$45\%$ band where multi-tool
  long-horizon failures dominate; small per-direction perturbations
  do not change which tasks succeed in nearly as many cases as on
  ALFWorld's $17$--$88\%$ range.
\end{itemize}
The right-side picture is the weakest: on Qwen2.5, the
parallel-on-right projection at $k{=}250$ recovered $71.8\%$ of the
oracle gain over base, evidence that $\dprl$'s input-feature scope
is largely shared with $\dwmrl$. On Qwen3, the analogous
parallel-on-right projection recovers only $46.9\%$ of the (much
smaller) oracle gain -- consistent with the
above noise/dynamic-range argument rather than with a failure of the
geometric claim, since the underlying right-overlap mean is in fact
higher on Qwen3 ($0.81$ vs $0.63$).

\paragraph{Takeaway.}
The U-disjoint / V-shared subspace asymmetry between $\dprl$ and
$\ddprl$, and the near-zero Frobenius cosine, are reproduced on a
different base model (Qwen3-8B vs Qwen2.5-7B), a different
post-training corpus ($\tau^2$-Bench vs ALFWorld), and a different
model depth ($36$ vs $28$ layers), with essentially identical left
numbers and an even sharper right alignment. The behavioral
projection sweep on Qwen3 carries the same qualitative sign
(orth $>$ par on the left at low $k$, no collapse under either
projection), but its margins are compressed into the noise band by
the smaller absolute $\dprl$ contribution and the harder benchmark.
Read together, these two appendices indicate that the geometric
complementarity diagnosed in \S\ref{sec:relationship} may be a property of the two-stage post-training recipe rather than of a particular
model/task, while the strength with which it shows up in
end-to-end success requires a task and model where $\dprl$ itself
has substantial headroom to begin with. However, we caution against over-interpreting these results: these experiments cannot establish generality across models and agent types.

\section{Preservation Method Details}
\label{app:preserve-details}

We study two ways to preserve the world-modeling ability acquired
during \wmrl while adapting the policy with \prl. The first is
\emph{training-free}: merge two fine-tuned checkpoints in weight
space. The second is \emph{training-based}: add an online
world-modeling SFT loss during \prl. In all cases, the preservation
method produces a single deployable policy checkpoint.

\paragraph{Training-free checkpoint merging.}
The merging methods take the base model $\mdl{0}$ and two task
checkpoints as input: always $\mdl{\wmrl}$ for task $A$, and either
$\mdl{\prl}$ or the sequential checkpoint $\mdl{\wmprl}$
($\wmrl\!\to\!\prl$) for task $B$. For each target weight matrix
$W^{(k)}$, we form task vectors
\begin{equation*}
\tau_i^{(k)} = W_i^{(k)} - W_0^{(k)}, \qquad i \in \{A,B\}.
\end{equation*}
We merge only the seven projection matrices in each transformer
layer,
\begin{equation*}
\mathtt{q,k,v,o,up,gate,down},
\end{equation*}
and leave embeddings, layer norms, and the lm-head at their
base-model values.

\emph{Plain TIES.}
Plain TIES operates directly on the raw task vectors
$\tau_A, \tau_B$ flattened from the entire model's parameters. For each pair of models, we stack the
two task vectors, keep the top-$K\%$ entries by magnitude
($K{=}40$), resolve sign conflicts with
\texttt{sum\_of\_values}, and average the surviving entries using a
disjoint mean, i.e. each coordinate is divided by the number of
non-zero contributors rather than by the total number of tasks. The
merged checkpoint is
\begin{equation*}
W_{\mathrm{merged}}
=
W_0
+
\mathrm{TIES}\!\left(\tau_A, \tau_B\right).
\end{equation*}

\emph{KnOTS.}
KnOTS first represents the two module-wise task vectors in a shared low-rank
SVD frame, then applies the same TIES masking and aggregation in
that frame. In the default $U$-sharing version, the two task
matrices are concatenated along the input dimension:
\begin{equation*}
\bigl[\tau_A^{(k)} \mid \tau_B^{(k)}\bigr]
=
U^{(k)} \Sigma^{(k)} V^{(k)\top}.
\end{equation*}
This gives a shared left basis $U^{(k)}$, corresponding to a shared
output subspace. The per-task right factors are split from
$\Sigma V^\top$, TIES is applied to those factors, and the merged
delta is reconstructed as
\begin{equation*}
\Delta^{(k)}
=
U^{(k)} V_{\mathrm{merged}}^{(k)}.
\end{equation*}

We also evaluate a $V$-sharing variant, motivated by the stronger
right-subspace overlap observed in \S\ref{sec:relationship}. This
variant transposes each task matrix before the joint SVD, so that
the shared basis lies on the input side rather than the output side.
After merging in this transposed frame, the reconstructed matrix is
transposed back to the original $(\mathit{out},\mathit{in})$ shape. We modify KnOTS's official repository \footnote{\url{https://github.com/gstoica27/knots}} to enable full parameter update merging.

For both KnOTS variants, we use the same TIES hyperparameters as
plain TIES: $K{=}40$, \texttt{sum\_of\_values} sign resolution,
disjoint-mean aggregation, scaling coefficient $1.0$, and no DARE
pre-processing. Singular directions with negligible singular values
are discarded.

\begin{algorithm}[t]
\caption{Online $\wmsft$ during \prl}
\label{alg:wmsft}
\begin{algorithmic}[1]
\REQUIRE Policy $\pi_\theta$, downsampling ratio $\rho=0.2$,
WM-SFT weight $\lambda=1.0$
\FOR{each \prl update step}
    \STATE Collect on-policy rollouts with $\pi_\theta$
    \STATE Extract real WM-SFT pairs
    $\mathcal{D}_{\mathrm{WM}}
    =
    \{((s_{t-H:t}, a_t), s_{t+1})\}$
    \STATE Downsample no-op transitions
    $(s_{t+1}=s_t)$ to at most $\rho$ times the number of
    changed-observation transitions
    \STATE Randomly keep $32$ samples each step
    \STATE Compute
    $\mathcal{L}_{\mathrm{GRPO}}$
    on the rollout batch
    \STATE Compute
    $\mathcal{L}_{\mathrm{WM\text{-}SFT}}$
    by predicting $s_{t+1}$ from $(s_{t-H:t}, a_t)$
    \STATE Update $\theta$ using
    $\mathcal{L}_{\mathrm{GRPO}}
    +
    \lambda\mathcal{L}_{\mathrm{WM\text{-}SFT}}$
\ENDFOR
\RETURN Updated policy $\pi_\theta$
\end{algorithmic}
\end{algorithm}

\paragraph{Training-based: online WM-SFT during \prl.}
We follow the \prl GRPO training setup of \citet{yu2026rwml}, but
add an auxiliary online world-modeling SFT update. At each training
step, we reuse the same on-policy rollouts collected for GRPO and
convert each realized transition into a next-observation prediction
example: given the recent observation history $s_{t-H:t}$ and the
executed action $a_t$, the model predicts the environment's next
observation $s_{t+1}$. Hence, the WM signal is ``free'' in the
sense of the concurrent ECHO recipe of \citet{shrivastava2026echo}:
it requires no additional environment interaction. The training
objective is
\begin{equation*}
\mathcal{L}
=
\mathcal{L}_{\mathrm{GRPO}}
+
\lambda
\mathcal{L}_{\mathrm{WM\text{-}SFT}},
\qquad
\lambda=1.0
\end{equation*}
with all other \prl hyperparameters unchanged.

\paragraph{KnOTS replication on $\tau^2$-Bench.}

\begin{table}[!htb]
\centering
\small
\setlength{\tabcolsep}{4pt}
\renewcommand{\arraystretch}{1.1}
\caption{KnOTS-TIES preservation interventions on Qwen3-8B +
$\tau^2$-Bench. We compare the source checkpoints with
training-free KnOTS-TIES merges at top-$K{=}40$, using either the
default shared $U$ basis or our shared $V$-basis variant. The two
merge blocks differ in the second task arm: \prl from base, or the
trained sequential $\wmrl\!\to\!\prl$ checkpoint. \textbf{Bold} =
best within each sub-block.}
\label{tab:qwen3-knots-merge}
\begin{tabular}{lccc}
\toprule
Method & AVG@3 & pass@1 & pass@3 \\
\midrule
\prl & $39.3 \pm 2.6$ & $38.0$ & $58.0$ \\
$\wmrl\!\to\!\prl$ & $45.0 \pm 1.6$ & $43.0$ & $64.0$ \\
\midrule
\multicolumn{4}{l}{\textit{training free: $\wmrl + \prl$}} \\
KnOTS-TIES ($U$) & $38.0 \pm 0.8$ & $37.0$ & $54.0$ \\
KnOTS-TIES ($V$) & $\mathbf{40.7 \pm 2.9}$ & $37.0$ & $\mathbf{58.0}$ \\
\midrule
\multicolumn{4}{l}{\textit{training free: $\wmrl + (\wmrl\!\to\!\prl)$}} \\
KnOTS-TIES ($U$) & $38.0 \pm 3.6$ & $35.0$ & $56.0$ \\
KnOTS-TIES ($V$) & $39.3 \pm 4.2$& $\mathbf{41.0}$ & $57.0$ \\
\bottomrule
\end{tabular}
\end{table}

We replicate the training-free KnOTS-TIES merges from
\S\ref{app:preserve-details} on Qwen3-8B + $\tau^2$-Bench, using the
same top-$K{=}40$, $\mathtt{sum\_of\_values}$ sign resolution, and
$\mathtt{merging\_type{=}mean}$ as in the ALFWorld experiments
(Table~\ref{tab:qwen3-knots-merge}). The results broadly favor the
shared right/input singular basis ($V$-sharing) over the default
$U$-sharing variant, consistent with the input-side geometry suggested
by \S\ref{sec:relationship}. In both merge settings, the $V$-sharing
variant attains the best AVG@3 within the training-free block, and in
the $\wmrl + (\wmrl\!\to\!\prl)$ setting it also improves pass@1 and
pass@3. However, the best training-free merge still remains below the
trained $\wmrl\!\to\!\prl$ sequential baseline, suggesting that simple
checkpoint merging is not sufficient to recover the full benefit of
staged training on this more complex benchmark. For cost reasons, we do
not replicate the online \wmsft auxiliary-loss baseline on
$\tau^2$-Bench.

\section{WMRL Data Example}
\label{app:wmrl_data_eg}
We show one representative training example from ALFWorld and one from
$\tau^2$-Bench. Each example consists of the model prompt, and the
ground-truth next observation used as the supervision target. Part of the content is omitted for display.

For $\wmsft$ and entropy probing in \Cref{fig:entropy}, we revise the final instruction to disable thinking.

\paragraph{ALFWorld example}
\label{app:wmrl_eg_alfworld}

\begin{examplebox}{Prompt}
\begin{PromptVerbatim}
You are an expert agent operating in the ALFRED Embodied Environment.
Your task is to: look at pillow under the desklamp.
Prior to this step, you have already taken 7 step(s). Below are the
most recent 2 observations and the corresponding actions you took:
[Observation 6: 'You arrive at shelf 1. On the shelf 1, you see a
 pencil 2, and a pencil 1.', Action 6: '[<action>go to desk 1</action>']
[Observation 7: 'Nothing happens.', Action 7: '<action>go to desk 1
 </action>']']

You are now at step 8 and your current observation is: Nothing
happens.

Your admissible actions of the current situation are:
['examine shelf 1' 'go to bed 1' 'go to desk 1' 'go to desk 2'
 'go to garbagecan 1' 'go to shelf 2' 'go to shelf 3' 'go to shelf 4'
 'go to shelf 5' 'inventory' 'look' 'take pencil 1 from shelf 1'
 'take pencil 2 from shelf 1'].

Potential action: <action>go to desk 1</action>

Now, your task is to predict the immediate next observation after executing the potential action above.
You should first *briefly* reflect on the previous steps and current situation. Then think about what the next observation will look like after taking this action, and also whether the task is completed after taking this action. This reflection and reasoning process MUST be enclosed within <think> </think> tags.
Once you've finished your reasoning, present both your final prediction of the next observation (use the past and current observations as examples!) and task completion status within <next_state> </next_state> tags.
\end{PromptVerbatim}
\end{examplebox}

\begin{examplebox}{Ground-truth next observation}
\begin{PromptVerbatim}
You arrive at desk 1. On the desk 1, you see a creditcard 1, and a
keychain 1. Task is not yet completed.
\end{PromptVerbatim}
\end{examplebox}

\begin{examplebox}{ALFWorld Non-thinking Prompt}
\begin{PromptVerbatim}
...
Now, your task is to predict the immediate next observation after executing the potential action above.
Directly present your final prediction of the next observation. DO NOT generate anything else.
\end{PromptVerbatim}
\end{examplebox}

\paragraph{$\tau^2$-Bench example}
\label{app:wmrl_eg_tau2}

\begin{examplebox}{Prompt, abbreviated}
\begin{PromptVerbatim}
<instructions>
You are a customer service agent that helps the user according to
the <policy> provided below.
...
</instructions>

<policy>
# Retail agent policy
[... full retail policy elided: ~6k chars covering cancel/modify
 pending orders, return/exchange delivered orders, identity
 authentication rules, payment-method rules, etc. ...]
</policy>

<tools>
[... ~30 tool schemas elided: find_user_id_by_email,
 find_user_id_by_name_zip, get_user_details, get_order_details,
 return_delivered_order_items, exchange_delivered_order_items,
 cancel_pending_order, modify_pending_order_items, transfer_to_human,
 ...]
</tools>

# Dialogue so far
user: Hi! I need to return two tablets I bought recently. Can you
 help me with that?
assistant: I can help you with that. To process your return, I'll
 need to verify your identity first.
...
potential assistant response: I've located your account, Chen Silva
(user ID: chen_silva_7485). To process your return, I'll need the
order ID of the tablets you purchased.

Once you've finished your thinking, format both your final prediction of the next user/tool response and task completion status within <next_state> </next_state> tags.

\end{PromptVerbatim}
\end{examplebox}

\begin{examplebox}{Ground-truth next observation}
\begin{PromptVerbatim}
Hmm, I don't have the order ID on me right now -- everything's kind of scattered. Is there another way we can find it? Maybe by the date I ordered or the tablet model? I think it was within the last two weeks.

Task is not yet completed.
\end{PromptVerbatim}
\end{examplebox}

\begin{examplebox}{Tau2-Bench Non-thinking Prompt}
\begin{PromptVerbatim}
...
DO NOT perform any thinking. Directly present your final prediction of the next user/tool response and task completion status.
\end{PromptVerbatim}
\end{examplebox}

\section{WM-SFT Prompt Format}
We use a lightweight next-observation prediction template for online $\wmsft$ to mitigate the collapse in the separate training stage.
The prompt contains the previous raw environment observation and the
agent's extracted action; the supervision target is only the next
environment observation.

\begin{examplebox}{Template}
\begin{PromptVerbatim}
Given the previous observation and the action, predict the next observation.

Previous observation:
{prev_obs}

Action:
{action}

Next observation:
\end{PromptVerbatim}
\end{examplebox}

\begin{examplebox}{Concrete instance}

\begin{PromptVerbatim}
<|im_start|>user
Given the previous observation and the action, predict the next observation.

Previous observation:
Nothing happens.

Action:
go to desk 1

Next observation:<|im_end|>
<|im_start|>assistant
\end{PromptVerbatim}
\end{examplebox}
Only these target tokens have \texttt{loss\_mask=1}
\begin{examplebox}{Supervision target}
\begin{PromptVerbatim}
You arrive at desk 1. On the desk 1, you see a creditcard 1, and a
keychain 1.<|im_end|>
\end{PromptVerbatim}
\end{examplebox}
\end{document}